\documentclass{article}

\usepackage{amsmath,amssymb,amsthm,mathtools}
\usepackage{bbm}

\usepackage{jmlr2e}

\usepackage{bm}
\usepackage{booktabs}
\usepackage{graphicx}
\usepackage{xcolor}
\usepackage{enumitem}
\usepackage{hyperref}
\usepackage{url}
\usepackage{thmtools}
\usepackage{dsfont}
\usepackage{makecell}

\hypersetup{
  colorlinks=true,
  linkcolor=blue,
  citecolor=blue,
  urlcolor=blue,
  pdfauthor={Atsushi Suzuki},
  pdftitle={Overfitting has a limitation: a model-independent generalization gap bound based on Rényi entropy}
}

\newcommand{\NewTerm}[1]{\textbf{\emph{#1}}}

\newcommand{\BM}[1]{\bm{\mathit{#1}}}

\newcommand{\Emph}[1]{\textbf{#1}}

\DeclareRobustCommand{\BM}[1]{\bm{#1}}
\DeclareMathOperator*{\argmin}{argmin}

\newcommand{\Renyi}{R\'{e}nyi}

\newcommand{\bb}[1]{\mathbb{#1}}
\newcommand{\calf}[1]{\mathcal{#1}}

\newcommand{\Risk}{\mathrm{Risk}}
\newcommand{\EmpRisk}{\mathrm{EmpRisk}}
\newcommand{\GenGap}{\mathrm{GenGap}}
\newcommand{\supp}{\mathrm{supp}}
\newcommand{\Uniform}{\mathrm{Uniform}}
\newcommand{\DI}{\mathrm{DI}}
\newcommand{\idop}{\mathrm{id}}

\newcommand{\EE}{\mathbb{E}}
\newcommand{\Prob}{\mathbb{P}}

\newcommand{\NN}{\mathbb{N}}
\newcommand{\RR}{\mathbb{R}}

\numberwithin{equation}{section}

\theoremstyle{plain}

\theoremstyle{definition}
\newtheorem{premise}{Premise}

\theoremstyle{remark}

\begin{document}

% --- Title / authors / abstract (from Typst show: neurips2025.with(...)) ---
\title{Overfitting has a limitation: a model-independent\\generalization gap bound based on \Renyi{} entropy}

\author{%
\name Atsushi Suzuki 
\email atsushi.suzuki.rd@outlook.com \\
\addr Department of Mathematics \\
Faculty of Science \\
The University of Hong Kong\\
Hong Kong SAR
\AND
\name Jing Wang 
\email jing.wang.research@gmail.com \\
\addr School of Computing and Mathematical Sciences \\
Faculty of Engineering and Science. \\
University of Greenwich\\
London, United Kingdom
}

\editor{My editor}

\maketitle

\begin{abstract}
Will further scaling up of machine learning models continue to bring success?
A significant challenge in answering this question lies in understanding generalization gap, which is the impact of overfitting.
Understanding generalization gap behavior of increasingly large-scale machine learning models remains a significant area of investigation, as conventional analyses often link error bounds to model complexity, failing to fully explain the success of extremely large architectures. This research introduces a novel perspective by establishing a model-independent upper bound for generalization gap applicable to algorithms whose outputs are determined solely by the data's histogram, such as empirical risk minimization or gradient-based methods. Crucially, this bound is shown to depend only on the R\'enyi entropy of the data-generating distribution, suggesting that a small generalization gap can be maintained even with arbitrarily large models, provided the data quantity is sufficient relative to this entropy. This framework offers a direct explanation for the phenomenon where generalization performance degrades significantly upon injecting random noise into data, where the performance degrade is attributed to the consequent increase in the data distribution's R\'enyi entropy. Furthermore, we adapt the no-free-lunch theorem to be data-distribution-dependent, demonstrating that an amount of data corresponding to the R\'enyi entropy is indeed essential for successful learning, thereby highlighting the tightness of our proposed generalization bound.
\end{abstract}


% =========================================================
\section{Introduction}
\label{sec:Introduction}

In fields such as natural language processing and video generation, machine learning (deep learning) using large-scale neural networks (NNs) with extremely high-dimensional parameters, trained on large-scale data, has recently achieved practical success in many areas (e.g., ChatGPT \citep{radford2019language,brown2020language,achiam2023gpt,hurst2024gpt}, Gemini \citep{team2023gemini,team2024gemini}, LLaMA \citep{touvron2023llama,dubey2024llama}, Claude \citep{anthropic2024claude}, Qwen \citep{bai2023qwen,yang2024qwen2,yang2024qwen25}, DeepSeek \citep{liu2024deepseek,guo2025deepseek}, Hunyuan \citep{sun2024hunyuan}, PaLM-E \citep{driess2023palm}, etc.).
Will machine learning continue to succeed by using extremely large machine learning models on even larger datasets in the future? This question can be rephrased as whether such extremely large models can reduce the \NewTerm{expected risk}, which is a quantification of the performance badness in machine learning. The expected risk is decomposed into the \NewTerm{empirical risk}, i.e., the loss on the training data, and the \NewTerm{generalization gap}, i.e., the difference between the expected risk and the empirical risk. When using large-scale machine learning models, it is not trivial whether the generalization gap can be made small.

Many existing analytical results suggest that the generalization gap worsens as the scale of the machine learning model increases. For example, although the worst-case generalization gap considering all hypotheses within the model can be evaluated using Rademacher complexity \citep{koltchinskii2000rademacher,koltchinskii2002empirical,lbartlett2002rademacher}, existing generalization gap analyses for NNs using Rademacher complexity depend on some measure of the NN's scale, such as the number of layers, the dimension of hidden layers, or the norm of weights \citep{neyshabur2015norm,bartlett2017spectrally,wei2019data,golowich2018sizeCOLT,golowich2020size,li2018tighter,harvey2017nearly,daniely2019generalization}. Therefore, they cannot explain the small generalization gap of extremely large models.
Generalization gap theories for cases where NNs can be compressed in some sense have also been studied within the Rademacher complexity framework \citep{arora2018stronger,suzuki2018spectral,Suzuki2020Compression} and the PAC-Bayes framework \citep{zhou2019non,lotfi2022pacbayes}, but they still depend on the scale of the NN, and it is not trivial under what circumstances NNs can be efficiently compressed.
Even when limiting the analysis to the generalization gap of the hypothesis that minimizes the empirical risk, theories such as the Akaike Information Criterion (AIC) \citep{akaike1974new} for regular models and the more general framework of local Rademacher complexity \citep{bartlett2005local,koltchinskii2006local} also assert that the generalization gap increases as the model becomes larger \citep{suzuki2018fast,terada2020fast}.
Generalization gap analyses for hypotheses selected by more practical optimization methods, such as stochastic gradient methods (e.g., \citep{cao2019generalization,jentzen2023overall}), also provide upper bounds on generalization gap that increase with the scale of the model, thus failing to explain the small generalization gap of ultra-large models. Moreover, it is not even guaranteed that ultra-large models used in the future will be constructed using NNs.

The above-mentioned upper bounds on generalization gap strongly depend on the model's construction. If a model-independent generalization gap theory could be developed, it would encourage the introduction of ultra-large models (which might include completely novel NN layers, or might not even be NNs). Such a model-independent theory is not impossible if we focus on the unevenness of a distribution. For example, if the true distribution of the data were concentrated at a single point, the generalization gap would be zero regardless of the machine learning model's construction. It has been known that in classification problems, the generalization gap on real data is small, whereas if the same model is applied to random labels, the empirical risk can be made small while the expected risk is naturally large, leading to an extremely large generalization gap, even when using the same model \citep{zhang2017understanding}. This cannot be explained in principle by focusing only on the model's construction. These observations suggest the necessity of focusing on the distribution in generalization gap analysis.

This paper shows for the first time that when using a machine learning algorithm whose hypothesis is determined by the histogram of the training data (a.k.a.\ a symmetric algorithm), such as training error minimization by exhaustive search or gradient methods, \Emph{there exists an upper bound on the generalization gap determined solely by the \Renyi{} entropy of the data-generating distribution}. Here, \Renyi{} entropy is a quantity that represents how far a probability distribution is from a uniform distribution. In other words, the theorem of this research asserts that if the data-generating distribution is uneven in the sense that it is far from a uniform distribution, the generalization gap will be small, regardless of the specific content or scale of the set of hypotheses included in the model.
Our bound does not depend on the set of hypotheses that the machine learning model comprises (e.g., the number of parameters, norms, etc.), or the property of the true hypothesis (the norm, sparseness, etc.). It can also apply to non-smooth and non-convex loss functions, as long as they are bounded.
Our generalization gap upper bound provides a clear and quantitative answer to the question of why the generalization gap increases when a part of the data is replaced with random numbers generated from a uniform distribution \citep{zhang2017understanding}, which could not be explained by existing theories that evaluate generalization gap by the size of the function space: it is because it increases the \Renyi{} entropy, on which the generalization gap depends exponentially. This is an advantage of the theory in this paper, which depends only on the distribution.
Furthermore, this research also shows that the sufficient condition for the length of training data for generalization, derived from the aforementioned generalization gap upper bound, is tight. Specifically, we extend the previously known no-free-lunch theorem for uniform distributions and show that, for non-uniform distributions, a data length of the order of the exponential of \Renyi{} entropy is necessary for successful learning.

The main contributions of this research are as follows:
\begin{enumerate}[label=(\arabic*)]
  \item We derived a novel generalization gap upper bound that depends only on \Renyi{} entropy, holding under the sole assumption that the algorithm is symmetric and independent of the specific construction of the machine learning model, and showed with a concrete example that it is not vacuous.
  \item We successfully explained the phenomenon where the generalization gap deteriorates by randomizing labels even when using the same machine learning model, from the perspective of an increase in \Renyi{} entropy.
  \item We derived a novel no-free-lunch theorem for non-uniform distributions, showing that the exponential of \Renyi{} entropy governs the data length required for learning, and that the aforementioned generalization gap upper bound is tight.
\end{enumerate}

\section{Related work} There is a large body of research on the generalization gap of large-scale machine learning models, especially NNs, particularly using Rademacher complexity (e.g., \citep{neyshabur2015norm,bartlett2017spectrally,wei2019data,golowich2018sizeCOLT,golowich2020size,li2018tighter,harvey2017nearly,daniely2019generalization,edelman2022inductive,gurevych2022on,takakura2023approximation,kim2024transformers}). As already mentioned, these depend strongly on the scale of the NN, while not using information about the distribution in the final generalization gap upper bound. Our theory, on the other hand, depends on the distribution but not on the scale of the model. A major technical difference is that the theory in this paper actively utilizes the fact that the data space is always a countable set. \Emph{Since machine learning is always implemented on computers, both the data space and the model are necessarily at most countable sets.} This paper actively uses this fact, which allows us to apply the method of types \citep{csiszar1982information}, leading us to non-tritival upper bounds. Upper bounds on generalization gap in PAC-Bayes theory \citep{mcallester1999pac} have also been derived for large-scale machine learning models \citep{zhou2019non,lotfi2022pacbayes}. This research also uses the countable hypothesis bound, a type of PAC-Bayes theory. Still, it differs significantly technically in that we use a prior distribution on the data space, which makes our theory model-independent, whereas conventional methods use a prior distribution on the parameter space, which causes the dependency on the size of the model. Some generalization gap analyses insist that they are model-independent \citep{chen2020dimension,roberts2021sgd}. However, they mean the \Emph{applicability} of their theories is model-independent, not the upper bounds' value.
For example, they depend on the norm of the true hypothesis parameter \citep{chen2020dimension}, the trace of the covariance matrix and the step size of the optimization algorithm, all of which tends to be large in large-scale machine learning models. In contrast, our upper bounds are model-independent in the sense of \Emph{their specific values}. Sibson's $\alpha$ mutual information, which is closely related to \Renyi{} entropy, has been used to derive generalization gap bounds (e.g., \citep{esposito2021generalization}). However, the bounds depend not on the data distribution but the mutual information between the training data and algorithm's outcome (similar to \citep{xu2017information,pensia2018generalization}), which can be vacuous when we effectively reduce the empirical error.
In contrast, our generalization gap bounds are valid and non-vacuous even when we completely minimize the objective function.

\noindent\textbf{The organization of the remainder of the paper} In Section~\ref{sec:Preliminary}, we provide the preliminaries for stating our results. Specifically, typical learning theory settings, symmetry of an algorithm, and \Renyi{} entropy are introduced.
In Section~\ref{sec:GenErr}, we provide our main theorem, a model-independent generalization gap bound determined by the data distribution \Renyi{} entropy. The section also explain the dependency of the generalization gap on data property demonstrated in previous work \citep{zhang2017understanding}. It also includes generalization gap bounds for specific scenarios.
Section~\ref{sec:NFL} provides a novel no-free-lunch theorem for non-uniform distributions, showing the tightness of our generalization gap upper bound.

% =========================================================
\section{Preliminaries}
\label{sec:Preliminary}

\paragraph{Notation.}
The set of all non-negative integers is denoted by $\NN$. Note that $0 \in \NN$.
The set of all real numbers is denoted by $\RR$.
When $\calf{X}$ and $\calf{Y}$ are sets, $\calf{X} \times \calf{Y}$ denotes the Cartesian product of $\calf{X}$ and $\calf{Y}$, and $\calf{Y}^{\calf{X}}$ denotes the set of all maps from $X$ to $Y$. That is,
\begin{equation}
  \calf{Y}^{\calf{X}} = \{ f \mid f : \calf{X} \to \calf{Y} \}.
\end{equation}
For a set $\calf{X}$ and $n \in \NN$, the Cartesian product of $n$ copies of $\calf{X}$ is denoted by $\calf{X}^n$.
When the generating distribution of a random variable $Z$ is ``$Q$'', it is written as $Z \sim Q$.
$Q^n$ denotes the $n$-fold product measure of $Q$.
That is, $Q^n$ is the distribution followed by a sequence of random variables $\BM{X} := (X_1,X_2,\dots,X_n)$ composed of $n$ independent random variables $X_1,X_2,\dots,X_n \sim Q$.
For a random variable $Z \sim Q$ on a set $\calf{Z}$ and a real-valued function $\phi : \calf{Z} \to \RR$ on $\calf{Z}$, the expected value of $\phi(Z)$ is written as $\EE_{Z \sim Q}\phi(Z)$. Also, the probability that an event $A(Z)$ depending on $Z$ occurs is written as $\Prob_{Z \sim Q}(A(Z))$. All logarithms in this paper are natural logarithms $\ln$.
\Emph{All the distributions to appear in this paper are discrete ones on an at most countable set}, since computers can handle those sets only. Hence, we identify probability mass functions with probability measures. That is, when a probability measure $Q$ on an at most countable set $\calf{A}$ is given, $Q(\{a\})$ for $a \in \calf{A}$ is simply written as $Q(a)$, and $Q$ is regarded as a probability mass function.

In the remainder of this section, we prepare for rigorously formulating the problem setting of this paper. Specifically, in Section \ref{ssec:Countable}, we introduce the premise that the data space is at most countable, which is maintained consistently throughout this paper, and explain why it always holds. In Section \ref{ssec:RiskDef}, we rigorously define the quantities we wish to evaluate in this paper, namely risks and generalization gaps, and intuitively explain why the at-most countability of the data space assumed in Section \ref{ssec:Countable} has a significant theoretical impact. In Section \ref{ssec:SymAlg}, we rigorously define symmetric algorithms, which are the subject of this paper. Examples include important and typical algorithms such as empirical risk minimization and gradient methods. In Section \ref{ssec:RenyiDef}, we introduce and rigorously define \Renyi{} entropy, which plays a central role in the evaluation of the generalization gap in this paper.

\subsection{Premise that the Data Space is At Most Countable: Why Does It Always Hold?} \label{ssec:Countable}

In this subsection, regarding the premise consistently placed in this paper that the data space is an at most countable set (i.e., a finite set or a countably infinite set), we explain why this is an assumption that can be unconditionally made in the context of machine learning using computers, and what significant impact this premise has on theory.

In this paper, we posit the following premise.

\begin{premise}
The data space is always an at most countable set (i.e., a finite set or a countably infinite set). That is, there always exists an injection from the data space to the set of natural numbers.
\end{premise}

In this paper, we call the above a premise rather than an assumption. This is because, as long as a computer is used, the above premise always holds. The reason why the above premise always holds as long as a computer is used is that the set consisting of all values that can be input into a computer is a countably infinite set. More specifically, any input to a computer is a finite binary sequence, but the set collecting all finite binary sequences $\{0, 1\}^* \coloneqq \bigcup_{L=0}^{+\infty} \{0, 1\}^L$ is a countably infinite set. As long as a computer is used, the data space is a subset of the set $\{0, 1\}^*$ collecting all such finite binary sequences, so it is at most countably infinite.

We have explained that the premise that the data space is an at most countable set can always be safely posited. Why this premise is theoretically important will be explained in Remark \ref{rem:CountabilityMatters} of Section \ref{ssec:RiskDef}.

% ---------------------------------------------
\subsection{Definition of Risk and Generalization Gap}
\label{ssec:RiskDef}

In this section, we rigorously define the quantities we wish to evaluate in this paper, namely risks and generalization gaps, and intuitively explain why the at-most countability of the data space assumed in Section \ref{ssec:Countable} has a significant theoretical impact.

\begin{definition}[Definition of Risk]
  \label{def:Risk}
  Let $\calf{Z}$ be a \Emph{countable} data space, $\calf{H}_{\mathrm{all}}$ be the whole hypothesis set, and $\ell : \calf{H}_{\mathrm{all}} \times \calf{Z} \to \bb{R}$ be a loss function defined on $\calf{Z}$ and $\calf{H}_{\mathrm{all}}$. Also, let $Q$ be a (discrete) probability measure on $\calf{Z}$, and consider a data sequence of length $n \in \NN$, $\BM{z} := (z_1,z_2,\dots,z_n) \in \calf{Z}^n$. At this time, the \NewTerm{expected risk function} $\Risk_{(\ell,Q)} : \calf{H}_{\mathrm{all}} \to \bb{R}$ on $Q$ and the \NewTerm{empirical risk function} $\EmpRisk_{(\ell,\BM{z})} : \calf{H}_{\mathrm{all}} \to \bb{R}$ on $\BM{z}$ are defined respectively as follows:
  \begin{equation}
    \Risk_{(\ell,Q)}(h) := \EE_{Z \sim Q} \ell(h,Z), \qquad
    \EmpRisk_{(\ell,\BM{z})}(h) := \frac{1}{n} \sum_{i=1}^n \ell(h,z_i).
  \end{equation}
  Furthermore, the \NewTerm{generalization gap function} $\GenGap_{(\ell,Q,\BM{z})} : \calf{H}_{\mathrm{all}} \to \bb{R}$ on $Q$ and $\BM{z}$ is defined by
  \begin{equation}
    \GenGap_{(\ell,Q,\BM{z})}(h) := \Risk_{(\ell,Q)}(h) - \EmpRisk_{(\ell,\BM{z})}(h).
  \end{equation}
  When clear from the context, $\ell$ is omitted.
\end{definition}

\begin{remark}[Meaning of Risks and Generalization Gap]
The loss $\ell(h,z)$ quantifies how bad the hypothesis $h \in \calf{H}_{\mathrm{all}}$ is on the data point $z \in \calf{Z}$.
Therefore, using the loss function $\ell$ and the true data generating distribution $Q$, the goal of machine learning can be formulated as finding $h \in \calf{H}_{\mathrm{all}}$ that minimizes the expected risk $\Risk_{(\ell,Q)}(h)$ as much as possible.
What is important is that the true data generating distribution $Q$ is unknown, so $\Risk_{(\ell,Q)}(h)$ cannot be directly calculated.
On the other hand, $\EmpRisk_{(\ell,\BM{z})}(h)$ can be calculated on the training data sequence $\BM{z} \in \calf{Z}^n$. Therefore, for the output $h$ of a machine learning algorithm, when the empirical risk $\EmpRisk_{(\ell,\BM{z})}(h)$ is calculated, we are interested in how much it differs from the expected risk $\Risk_{(\ell,Q)}(h)$, i.e., the generalization gap $\GenGap_{(\ell,Q,\BM{z})}(h)$. This is why the evaluation of generalization gap is important in the field of machine learning. The phenomenon where the generalization gap becomes large is called overfitting.
\end{remark}

Since the above definition is somewhat abstract, let us look at how actual problems fit into the above definition, using a classification problem as an example. Note that the definition of a classification problem below is broad, so please be aware that it includes not only classical binary classification but also practical Chat AI.

\begin{example}[Classification Problem]
In the case of a classification problem, the data space is given by the Cartesian product of the input data space $\calf{X}$ and the output data space $\calf{Y}$, i.e., $\calf{Z} = \calf{X} \times \calf{Y}$. For a deterministic classification problem, the whole hypothesis set is the set of all maps from $\calf{X}$ to $\calf{Y}$, i.e., $\calf{H}_{\mathrm{all}} = \calf{Y}^{\calf{X}}$. Then, the 0-1 loss
\begin{equation}
  \ell_\textrm{0-1} : \calf{Y}^{\calf{X}} \times (\calf{X} \times \calf{Y}) \to \RR
\end{equation}
is defined as
\begin{equation}
  \ell_\textrm{0-1}(f,(x,y)) := \mathds{1}(y \neq f(x)) :=
  \begin{cases}
    1 & \text{if } y \neq f(x),\\
    0 & \text{otherwise,}
  \end{cases}
\end{equation}
where $f \in \calf{Y}^{\calf{X}}$, $x \in \calf{X}$, and $y \in \calf{Y}$.
At this time, the expected risk of $f \in \calf{H}_{\mathrm{all}} = \calf{Y}^{\calf{X}}$ is
\begin{equation}
  \Risk_{(\ell,Q)}(f)
  = \EE_{Z \sim Q} \ell(f,Z)
  = \Prob_{(X,Y)\sim Q}(Y \neq f(X)),
\end{equation}
which is the misclassification rate of $f$ in the true distribution, so this is exactly what we want to minimize in a classification problem. Considering a natural language chatbot (Chat AI), both the input data set and the output data set can be infinite. They are sets of finite-length strings
\begin{equation}
  \Sigma^{*} := \Sigma^{0} \cup \Sigma^{1} \cup \Sigma^{2} \dotsc,
\end{equation}
Here, $\Sigma$ is a character set (e.g., all ASCII characters) and is a finite set, and for $l \in \NN$, $\Sigma^{l}$ is the set of all strings of length $l$.
\end{example}

\begin{remark}[Essential theoretical differences between continuous probability distributions on real vector spaces and discrete probability distributions on at most countable sets]
\label{rem:CountabilityMatters}
In many contexts of learning theory \citep{koltchinskii2000rademacher,koltchinskii2002empirical,lbartlett2002rademacher,neyshabur2015norm,bartlett2017spectrally,wei2019data,golowich2018sizeCOLT,golowich2020size,li2018tighter,harvey2017nearly,daniely2019generalization}, the data space is a subset of a real vector space, and the data generating distribution is often assumed to be a continuous probability distribution on that set. That is, it was common to assume a situation where a probability density function $q$ exists, and for any Lebesgue measurable set $A$, the probability $\mathop{\mathrm{Pr}}(z \in A)$ that the generated data $z \in \mathbb{R}^d$ ($d$ is the dimension of the data space) is contained in the set $A$ is always given by the integral of $q$ over $A$:
\begin{equation}
  \mathop{\mathrm{Pr}}(z \in A) = \int \mathbbm{1} \{z \in A\} q (z) d z,
\end{equation}
where $\mathbbm{1} \{z \in A\}$ is the characteristic function that returns 1 if $z \in A$ and 0 otherwise.
What is notable about this situation of continuous probability distributions is that, with probability 1, the training data sequence determining the empirical risk and the newly generated data determining the expected risk become ``unrelated.'' Specifically, when considering a training data sequence $(z_1, z_2, ..., z_n)$ of length $n$ generated from the probability distribution defined by the probability density function $q$, and a random variable $Z$ generated from the probability distribution also defined by $q$, the probability that $Z \in \{z_1, z_2, ..., z_n\}$ is zero. This fact is evident from the property that the measure of a finite set is always zero in continuous probability distributions; this means that, with probability 1, the set formed by the training data sequence and the support of the singleton set consisting of the newly generated data do not overlap. This implies that, on continuous probability distributions, no relationship can be expected between the expected risk and the empirical risk without additional assumptions regarding the loss function. Therefore, to obtain a meaningful generalization gap, assumptions that restrict the relationship of loss values at different data points, such as Lipschitz continuity of the loss function, are necessary.

In contrast, this paper takes as a premise that the data space is an at most countable set. For a probability distribution on an at most countable data space, a probability mass function is always defined. When considering a training data sequence $(z_1, z_2, ..., z_n)$ of length $n$ generated from the probability distribution defined by a certain probability mass function $Q$, and a random variable $Z$ generated from the probability distribution also defined by $Q$, the probability that $Z \in \{z_1, z_2, ..., z_n\}$ is always positive. Therefore, as long as the loss function is a function uniquely determined by the input, the expected risk and the empirical risk cannot be completely unrelated. Consequently, there is a possibility of obtaining a meaningful generalization gap without constraints on the function class.
\end{remark}

Machine learning referred to the process in which a machine determines a single hypothesis from a data sequence using a learning algorithm. Let us strictly define the learning algorithm below.

\begin{definition}[Model and Learning Algorithm]
A subset $\calf{H} \subset \calf{H}_{\mathrm{all}}$ of the whole hypothesis set is called a model.
A map
\begin{equation}
  \mathfrak{A} : \calf{Z}^{*} \to \calf{H},
\end{equation}
from the set of finite data sequences
\begin{equation}
  \calf{Z}^{*} := \calf{Z}^{0} \cup \calf{Z}^{1} \cup \calf{Z}^{2} \cup \dotsc
\end{equation}
to the model $\calf{H}$ is called a learning algorithm.
\end{definition}

% ---------------------------------------------
\subsection{Symmetry of Algorithms}
\label{ssec:SymAlg}

The empirical risk function is determined by the histogram of the data sequence and does not depend on the order of appearance of each data point. In other words, in a typical setting, we are not interested in the order of the training data points. This means that information about the order of the data sequence can be disregarded. Therefore, when considering algorithms, it is natural to consider algorithms whose output is determined by the histogram of the data sequence and does not depend on the order of appearance of each data point. Such algorithms are called symmetric algorithms (e.g., \citep{nikolakakis2022beyond}).

Below, we start by defining symmetric maps more generally.

\begin{definition}[Symmetry of a Map]
  \label{def:SymMap}
  For a data space $\calf{Z}$, a map $\phi : \calf{Z}^{*} \to \calf{T}$
  from the set of finite data sequences $\calf{Z}^{*} := \calf{Z}^{0} \cup \calf{Z}^{1} \cup \calf{Z}^{2} \cup \dotsc$
  to some set $\calf{T}$ is symmetric if, for any permutation $\sigma \in \mathfrak{S}_{n}$ of $n$ elements,
  \begin{equation}
    \phi(z_1,z_2,\dots,z_n) = \phi(z_{\sigma(1)},z_{\sigma(2)},\dots,z_{\sigma(n)})
  \end{equation}
  holds.

  In other words, $\phi$ is symmetric means that $\phi(\BM{z})$ is determined solely by the histogram of $\BM{z}$ and does not depend on the order of appearance of the data.
\end{definition}

\begin{example}[Important symmetric maps in machine learning]
\begin{enumerate}[label=(\arabic*)]
  \item \textit{Empirical risk function}: When a hypothesis $h \in \calf{H}_{\mathrm{all}}$ is fixed, the empirical risk considered as a function of the data sequence, $\EmpRisk_{(\ell,\cdot)}(h) : \calf{Z}^{*} \to \RR$,
  is a symmetric map (real-valued function). This can be seen from the fact that the empirical risk depends only on the histogram of the data, not on its order.
  \item \textit{Gradient of empirical risk}: When hypotheses are identified with elements of a real vector space (i.e., parameterized by real vectors), its gradient in that real vector space, $\nabla \EmpRisk_{(\ell,\BM{z})}(h)$,
  is a symmetric map (real vector-valued function).
\end{enumerate}
\end{example}

\begin{definition}[Symmetry of a Learning Algorithm]
  \label{def:SymAlg}
  A learning algorithm $\mathfrak{A} : \calf{Z}^{*} \to \calf{H} \subset \calf{H}_{\mathrm{all}}$
  is symmetric if $\mathfrak{A}$ is symmetric as a map in the sense of Definition~\ref{def:SymMap}.
\end{definition}

\begin{example}[Examples of Symmetric Learning Algorithms]
As a simple observation, if each step of a learning algorithm depends on the data sequence only through symmetric functions, then the learning algorithm is symmetric. Important examples are listed below.

\begin{enumerate}[label=(\arabic*)]
  \item \textit{Empirical risk minimization by exhaustive search}: This can be written as
  \begin{equation}
    \mathfrak{A}(\BM{z}) = \argmin_{h \in \calf{H}} \EmpRisk_{(\BM{z})}(h).
  \end{equation}
  The fact that this empirical risk minimization is a symmetric learning algorithm follows from the fact that the empirical risk is a symmetric function with respect to the data sequence.
  \item \textit{Gradient method with a fixed initial point}: This is a general term for methods where the initial hypothesis is $h_0 \in \calf{H}$, the hypothesis $h_t \in \calf{H}$ at step $t$ is selected depending on the history of past empirical risk gradients $(\nabla \EmpRisk_{(\ell,\BM{z})}(h_\tau))_{\tau=0}^{t-1}$ and the history of past selected hypotheses $(h_\tau)_{\tau=0}^{t-1}$, and the stopping condition also depends only on these. Note that this formulation allows the use of gradient information for $\tau < t-1$, so it includes algorithms that use auxiliary variables in practice (e.g., Nesterov's accelerated gradient method \citep{nesterov1983method}, BFGS method \citep{fletcher1970new,goldfarb1970family,shanno1970conditioning}). Gradient methods are symmetric learning algorithms because the gradient of the empirical risk is a symmetric (real vector-valued) map with respect to the data sequence.
\end{enumerate}
\end{example}

\begin{remark}[Discussing stochastic symmetric algorithms is important future work]
  \label{rem:FutureStochastic}
  In this paper, we only consider \Emph{deterministic} symmetric methods, but do not consider stochastic symmetric algorithms.
  Since stochastic symmetric algorithms include algorithms widely used in modern machine learning, including stochastic gradient descent and Adam \citep{kingma2014adam}, extending this paper's discussion to those algorithms is important future work.
\end{remark}

% ---------------------------------------------
\subsection{\Renyi{} Entropy as a Measure of Distribution Unevenness}
\label{ssec:RenyiDef}

The unevenness of a distribution has a large impact on generalization gap. To give an extreme example, no matter how large-scale a machine learning model is used, if the data distribution degenerates to a single point, the generalization gap is zero. Even if not so extreme, there is an intuition that if the data is skewed, the generalization gap will be small. As an example, as already mentioned, even in practical deep learning models, there are known cases where replacing part of the data with uniform random numbers causes a sharp increase in generalization gap \citep{zhang2017understanding}.
This section introduces \Renyi{} entropy as an indicator to quantify the unevenness of a distribution.

\begin{definition}
Let $\alpha \in [0,+\infty]$.
The $\alpha$-\NewTerm{\Renyi{} entropy} $H_\alpha(Q) \in [0,+\infty]$ of a discrete probability distribution $Q$ defined on an at most countable set $\calf{Z}$ is defined as follows:
\begin{equation}
H_\alpha(Q) =
\begin{cases}
  \displaystyle \sum_{z \in \calf{Z}} Q(z)\, \ln \frac{1}{Q(z)}, & \text{if } \alpha = 1,\\[1.2ex]
  \displaystyle \ln \bigl|\supp(Q)\bigr|, & \text{if } \alpha = 0,\\[1.2ex]
  \displaystyle - \ln \left(\max_{z \in \calf{Z}} Q(z) \right), & \text{if } \alpha = \infty,\\[1.2ex]
  \displaystyle \frac{1}{1-\alpha} \ln \left( \sum_{z \in \calf{Z}} Q(z)^{\alpha} \right), & \text{otherwise,}
\end{cases}
\end{equation}
where $\supp(Q) := \{ z \in \calf{Z} \mid Q(z) > 0 \}$.
\end{definition}

\begin{remark}[Meaning of \Renyi{} Entropy]
  \label{rem:Renyi}
  $H_\alpha(Q)$ represents, in some sense, the ``unevenness'' or ``effective support size'' (logarithm thereof) of the distribution $Q$. This can also be understood from the following observations:
  \begin{enumerate}[label=(\arabic*)]
    \item For any fixed $\alpha \in [0,+\infty]$, $H_\alpha(Q)$ takes its minimum value of $0$ if and only if $Q$ is a point measure (i.e., $\exists z \in \calf{Z}, Q(z) = 1$).
    \item If the support set $\calf{Z}$ is finite, then for any fixed $\alpha \in [0,+\infty]$, $H_\alpha(Q)$ takes its maximum value $\log |\calf{Z}|$ if and only if $Q$ is a uniform distribution on $\calf{Z}$.
  \end{enumerate}
  Note that, for a fixed probability distribution $Q$, $H_\alpha(Q)$ is continuous and monotonically non-increasing with respect to $\alpha$. This is because as $\alpha$ increases, the weights of elements with small probability mass are reduced, effectively ignoring them.
\end{remark}

% =========================================================
\section{Generalization Gap Bound Determined by \Renyi{} Entropy}
\label{sec:GenErr}

This section presents the quantitative relation among the generalization gap and \Renyi{} entropy and the training data size. 
Section \ref{ssec:GapFromData} provides generalization gap upper bound determined by the \Renyi{} entropy and the training data length.
By solving the inequality with respect to the training data length, Section \ref{ssec:DataFromGap} provides a sufficient condition with respect to the training data length to keep the generalization gap lower than the given aimed standard.
Based on the inequalities provided by those sections, Section \ref{ssec:GapFromNoise} explains why mixing noise worsens the generalization gap.
Section \ref{sub:Specific} provides specific formulae of the generalization gap upper bound and the sufficient condition with respect to the training data size for typical distributions.

\subsection{Main Theorem: Generalization Gap Bound Theorem Determined by \Renyi{} Entropy}
\label{ssec:GapFromData}

This section presents the main theorem of this paper, the generalization gap bound theorem determined by \Renyi{} entropy. The following is the main theorem, which gives a probabilistic upper bound on the generalization gap when the number of data points is fixed.

\begin{theorem}[Generalization Gap Bound by \Renyi{} Entropy]
  \label{thm:RenyiGenErr}
  Fix a whole hypothesis set $\calf{H}_{\mathrm{all}}$ and a loss function $\ell : \calf{H}_{\mathrm{all}} \times \calf{Z} \to \RR$ defined on a data space $\calf{Z}$ which is an at most countable set.
  Define
  \begin{equation}
    \DI(\ell) := \sup_{h \in \calf{H}_{\mathrm{all}},\, z \in \calf{Z}} \ell(h,z)
    - \inf_{h \in \calf{H}_{\mathrm{all}},\, z \in \calf{Z}} \ell(h,z) \in [0,+\infty]
  \end{equation}
  (``$\DI$'' means the diameter of the image). Let $\calf{Z}^{*} := \calf{Z}^{0} \cup \calf{Z}^{1} \cup \calf{Z}^{2} \cup \dotsc$
  be the set of all finite-length data sequences, and let
  $\mathfrak{A} : \calf{Z}^{*} \to \calf{H}_{\mathrm{all}}$
  be a symmetric machine learning algorithm in the sense of Definition~\ref{def:SymAlg}. Let $Q$ be a probability distribution on $\calf{Z}$, and for $\alpha \in [0,1]$, define $\kappa_{(Q,\alpha)} : \NN \to \RR$ by
  \begin{equation}
    \kappa_{(Q,\alpha)}(n) := n^{\alpha} \exp\left((1-\alpha) H_\alpha(Q)\right),
  \end{equation}
  and define $\kappa^{*}_{(Q)} : \NN \to \RR$
  by 
  \begin{equation}
    \kappa^{*}_{(Q)}(n) := \min_{\alpha \in [0,1]} \kappa_{(Q,\alpha)}(n).
  \end{equation}

  When $n \in \NN_{>0}$ and $\BM{Z} = (Z_1,Z_2,\dots,Z_n) \sim Q^{n}$, i.e., $Z_1,Z_2,\dots,Z_n \sim Q$ independently, for any $\delta_1,\delta_2,\delta_3 > 0$, the following holds with probability at least $1 - (\delta_1 + \delta_2 + \delta_3)$:
  \begin{equation}
    \GenGap_{(\ell,Q,\BM{Z})}\left(\mathfrak{A}(\BM{Z})\right) \leq
    \DI(\ell)
    \sqrt{
      \frac{
      \left(\kappa^{*}_{(Q)}(n) + \sqrt{ \tfrac{n}{2} \ln \tfrac{2}{\delta_3} } \right)
      \left( 3 \ln n + \ln(2\pi) + \ln \tfrac{1}{\delta_2} \right)
      + \ln \tfrac{1}{\delta_1}
      }{2n}
    }.
  \end{equation}
\end{theorem}

\begin{remark}[Theorem~\ref{thm:RenyiGenErr} is model-independent.]
  \label{rem:ModelIndependent}
  Theorem~\ref{thm:RenyiGenErr} holds regardless of the construction of each hypothesis $h$, the structure of the hypothesis set $\calf{H}$, or the relationship between the hypothesis and the loss function $\ell$. No matter how complex a function an individual $h$ is, no matter how many parameters $\calf{H}$ has or how complex a model it is constructed with, and no matter how discontinuously $\ell$ behaves with respect to $h$ or $z$, Theorem~\ref{thm:RenyiGenErr} holds. In that sense, Theorem~\ref{thm:RenyiGenErr} is model-independent.
\end{remark}

\begin{remark}[Rough behavior of the generalization gap upper bound]
  \label{rem:RoughBehavior}
  Let
  \begin{equation}
    \alpha^{*} := \argmin_{\alpha \in [0,1]} \kappa_{(Q,\alpha)}(n).
  \end{equation}
  In this case, the upper bound of Theorem~\ref{thm:RenyiGenErr} is roughly of the order $\sqrt{ n^{\alpha^{*}-1} \exp\left( (1-\alpha^{*}) H_{\alpha^{*}}(Q) \right) }$
  If we ignore the dependence of $\alpha^{*}$ on $H_\alpha(Q)$, the upper bound is exponential w.r.t.\ the \Renyi{} entropy.
  A more detailed discussion will be provided later.
\end{remark}

\begin{remark}[Trade-off regarding $\alpha$]
  \label{rem:TradeoffAlpha}
  To minimize the right-hand side, one should minimize
  \(
    n^\alpha \exp\left((1-\alpha)H_\alpha(Q)\right)
  \)
  with respect to $\alpha$.
  Since \Renyi{} entropy is a non-increasing function of $\alpha$, $\exp\left((1-\alpha)H_\alpha(Q)\right)$ is a decreasing function in the range $\alpha \in [0,1]$. On the other hand, $n^\alpha$ is an increasing function of $\alpha$. To obtain a good upper bound, it is necessary to determine a good $\alpha$ within this trade-off.
  As an extreme case, if we consider $\alpha=1$, then
  \(
    n^\alpha \exp\left((1-\alpha)H_\alpha(Q)\right) = n.
  \)
  In this case, the right-hand side becomes $O(\ln n)$, which is a vacuous bound that does not converge to $0$ even if $n$ is increased. Therefore, an appropriate choice of $\alpha$ is essential.
\end{remark}

\begin{remark}[Case where \Renyi{} entropy diverges]
  \label{rem:InfiniteEntropy}
  There exist distributions $Q$ for which \Renyi{} entropy $H_\alpha(Q)$ always diverges in the range $\alpha \in [0,1]$.
  This is equivalent to the divergence of Shannon entropy $H_1(Q)$.
  For example, a probability distribution on $\calf{Z} = \NN$ with
  \begin{equation}
    Q(k) := \frac{1}{C (k+2)(\ln(k+2))^{2}},
  \end{equation}
  where
  \begin{equation}
    C := \sum_{k' = 0}^{+\infty} \frac{1}{(k'+2)(\ln(k'+2))^{2}} < +\infty,
  \end{equation}
  is such an example.
  If \Renyi{} entropy $H_\alpha(Q)$ always diverges in the range $\alpha \in [0,1]$, the upper bound of Theorem~\ref{thm:RenyiGenErr} is vacuous.
  However, this is a pathological case, and in such cases, as will be discussed later, it includes cases where learning from finite-length training data is known to be impossible in the sense of the no-free-lunch theorem.
  Also, as will be discussed later, the upper bound of Theorem~\ref{thm:RenyiGenErr} is usually not vacuous even when the tail probability of $Q$ decays according to a power law.
\end{remark}

% ---------------------------------------------
\subsection{Sufficient Data Length for Small Generalization Gap}
\label{ssec:DataFromGap}

The previous section provided an upper bound on the generalization gap when the data length is fixed. Conversely, we are often interested in the sufficient condition for the data length to achieve a target generalization gap. Essentially, this involves solving Theorem~\ref{thm:RenyiGenErr} for $n$, but expressing the sufficient condition for data length using elementary functions is a somewhat tedious task because it involves the inverse function of a product of a polynomial and a logarithmic function. This section presents the result of that task and shows that \Renyi{} entropy, i.e., the unevenness of the data distribution, has a significant impact on the sufficient condition for data length to achieve small generalization gap. The following theorem is the mathematical statement.

\begin{theorem}[Sufficient condition for data length determined by \Renyi{} entropy]
  \label{thm:GeneralDataLength}
  Assume the same situation as in Theorem~\ref{thm:RenyiGenErr}.
  That is, fix a whole hypothesis set $\calf{H}_{\mathrm{all}}$, a loss function $\ell : \calf{H}_{\mathrm{all}} \times \calf{Z} \to \RR$ defined on a data space $\calf{Z}$ which is an at most countable set, define $\DI(\ell)$ similarly, and let $\mathfrak{A} : \calf{Z}^{*} \to \calf{H}_{\mathrm{all}}$ be a symmetric machine learning algorithm in the sense of Definition~\ref{def:SymAlg}.

  For a (discrete) probability measure $Q$ on $\calf{Z}$, define the extended real-valued functions $\nu_{(Q,\alpha)} : (0,1] \to [0,+\infty)$ and $\widetilde{\nu}_{(Q,\alpha)} : (0,1]^{2} \to [0,+\infty)$
  by
  \begin{equation}
    \nu_{(Q,\alpha)}(\varepsilon) :=
    \left( \frac{24 H_\alpha(Q) \ln \frac{12}{\varepsilon^{2}(1-\alpha)}}{\varepsilon^{2}} \right)^{\frac{1}{1-\alpha}}
    \exp\left( H_\alpha(Q) \right),
  \end{equation}
  \begin{equation}
    \widetilde{\nu}_{(Q,\alpha)}(\delta,\varepsilon) :=
    \left( \frac{36 \ln \frac{6\pi}{\delta}}{\varepsilon^{2}} \right)^{\frac{1}{1-\alpha}}
    \exp\left( H_\alpha(Q) \right).
  \end{equation}

  Also, define $\omega : (0,1]^{2} \to [0,+\infty)$
  by
  \begin{equation}
    \omega(\delta,\varepsilon)
    =
    \max \left\{
      \frac{324 \ln \frac{3}{\delta}}{\varepsilon^{4}}
      \left(
        \ln \frac{9 \sqrt{2 \ln \frac{3}{\delta}}}{\varepsilon^{2}}
      \right)^{2},
      \,
      \frac{3}{2 \varepsilon^{2}} \ln \frac{3}{\delta}
    \right\}.
  \end{equation}

  Fix any $(\delta,\varepsilon) \in (0,1)^{2}$. If for some $\alpha \in [0,1]$,
  \begin{equation}
    n > \max \Bigl\{ \nu_{(Q,\alpha)}(\varepsilon), \widetilde{\nu}_{(Q,\alpha)}(\delta,\varepsilon), \omega(\delta,\varepsilon) \Bigr\}
  \end{equation}
  holds, then, when $\BM{Z} = (Z_1,Z_2,\dots,Z_n) \sim Q^{n}$, with probability at least $1-\delta$,
  \begin{equation}
    \GenGap_{(\ell,Q,\BM{Z})} \left( \mathfrak{A}(\BM{Z}) \right) < \DI(\ell)\, \varepsilon.
  \end{equation}
\end{theorem}

\begin{remark}[Theorem~\ref{thm:GeneralDataLength} is also model-independent.]
  \label{rem:DataLengthModelIndependent}
  Theorem~\ref{thm:GeneralDataLength} is model-independent in the same sense as stated in Remark~\ref{rem:ModelIndependent}.
\end{remark}

\begin{remark}[The main term is $\nu_{(Q,\alpha)}$]
  \label{rem:DataLengthRenyiRelation}
  In large-scale problems, $H_\alpha(Q)$ usually becomes large, but $\omega(\delta,\varepsilon)$ does not depend on $H_\alpha(Q)$. Also, if $H_\alpha(Q) \gg \ln\frac{1}{\delta}$, then $\nu_{(Q,\alpha)}(\varepsilon) \gg \widetilde{\nu}_{(Q,\alpha)}(\delta,\varepsilon)$. Therefore, $\nu_{(Q,\alpha)}$ is the main term. The specific form of $\nu_{(Q,\alpha)}$ implies that the data length should be at least proportional to $\exp(H_\alpha(Q))$, the exponential of the \Renyi{} entropy.
\end{remark}

% ---------------------------------------------
\subsection{Why does mixing uniform random noise into data worsen generalization gap?}
\label{ssec:GapFromNoise}

It is known that deep learning models used in practical image recognition have low generalization gap on original data (both training error rate and test error rate are low), but if the data labels are randomized, the generalization gap becomes extremely large (training error rate is low, but test error rate is high) \citep{zhang2017understanding}.
This phenomenon cannot be explained in principle by theories that focus only on the function class represented by the model. This section provides a direct explanation for this phenomenon from the perspective of an increase in \Renyi{} entropy. More specifically, replacing a part of the data with uniform random numbers increases the \Renyi{} entropy. Quantitatively, the following holds.

\begin{proposition}[Deterministic label vs uniform random label]
  \label{prp:labels}
  Let a random variable $X$ on $\calf{X}$ follow a probability distribution $Q$. Let a random variable $Y$ on a finite set $\calf{Y}$ be given by $Y = f(X)$ using a deterministic function $f$. Let a random variable $Y'$ on $\calf{Y}$ follow a uniform distribution on $\calf{Y}$ independently of $X$. Then, for any $\alpha \in [0,+\infty]$,
  \begin{equation}
    H_\alpha(X,Y') = H_\alpha(X,Y) + \ln |\calf{Y}|.
  \end{equation}
\end{proposition}

\begin{proof}
Since the probability distribution of $(X,Y)$ is effectively the same as the probability distribution of $X$, $H_\alpha(X,Y) = H_\alpha(X) = H_\alpha(Q)$. Also, from the independence of $X$ and $Y'$, it follows that $H_\alpha(X,Y') = H_\alpha(X) + H_\alpha(Y')$. Since $Y'$ follows a uniform distribution, $\forall \alpha \in [0,+\infty], H_\alpha(Y') = \ln |\calf{Y}|$.
\end{proof}

When \Renyi{} entropy increases additively, there is an exponential effect on the generalization gap.

\begin{theorem}[Deterioration of generalization gap caused by an increase in \Renyi{} entropy]
  \label{thm:RenyiCauseGenErr}
  Suppose that for two probability distributions $Q$ and $Q'$, there exists some $C \ge 0$ such that
  \begin{equation}
    \forall \alpha \in [0,1], \quad H_\alpha(Q') \ge H_\alpha(Q) + C.
  \end{equation}
  Then, for any $n \in \NN$,
  \begin{equation}
    \kappa^{*}_{(Q')}(n) \ge \exp\left( (1-\alpha'^{*}) C \right) \, \kappa^{*}_{(Q)}(n),
  \end{equation}
  where
  \begin{equation}
    \alpha'^{*} := \argmin_{\alpha \in [0,1]}
    \exp\left( (1-\alpha) H_\alpha(Q') \right) n^{\alpha}.
  \end{equation}
  Also, for any $\alpha \in [0,1]$ and any $(\delta,\varepsilon) \in (0,1]^{2}$,
  \begin{equation}
    \max\bigl\{ \nu_{(Q',\alpha)}(\varepsilon),\, \widetilde{\nu}_{(Q',\alpha)}(\delta,\varepsilon) \bigr\}
    \ge
    \exp(C)\, \max\bigl\{ \nu_{(Q,\alpha)}(\varepsilon),\, \widetilde{\nu}_{(Q,\alpha)}(\delta,\varepsilon) \bigr\}.
  \end{equation}
\end{theorem}

\begin{remark}[Deterioration of generalization gap can be explained by the increase in \Renyi{} entropy]
  \label{rem:RenyiCauseGenErr}
  According to Theorem~\ref{thm:RenyiGenErr}, the main term of the upper bound on generalization gap was $O\left( \sqrt{ \kappa^{*}_{(Q)}(n) / n \cdot \ln n } \right)$. Therefore, the generalization gap for the probability distribution $Q'$ is roughly $\sqrt{\exp(C)^{1-\alpha^{*}}}$ times worse than for $Q$. Considering the example in Proposition~\ref{prp:labels}, $C = \ln |\calf{Y}|$, so in the case of uniform labels $(X,Y')$, the generalization gap is $\sqrt{|\calf{Y}|^{\,1-\alpha^{*}}}$ times worse than in the case of deterministic labels $(X,Y)$.
  When $\alpha'^{*} = 1$, the inequality
  \(
    \kappa^{*}_{(Q')}(n) \ge \exp\left( (1-\alpha'^{*})C \right) \kappa^{*}_{(Q)}(n)
  \)
  is meaningless, but such cases are when Theorem~\ref{thm:RenyiGenErr} gives a vacuous upper bound, and as we will see in a later example, such cases are rare.

  Furthermore, the sufficient data length $n$ to make the generalization gap less than or equal to $\DI(\ell) \varepsilon$ is effectively given by
  \(
    \max\bigl\{ \nu_{(Q',\alpha)}(\varepsilon), \widetilde{\nu}_{(Q',\alpha)}(\delta,\varepsilon) \bigr\}.
  \)
  Therefore, applying the above theorem, the sufficient condition for data length in the case of probability distribution $Q'$ is $\exp(C)$ times worse than for $Q$. Considering the example in Proposition~\ref{prp:labels} again, in the case of uniform labels $(X,Y')$, the sufficient condition for data length is $|\calf{Y}|$ times worse than in the case of deterministic labels $(X,Y)$. This is why the generalization gap deteriored when the labels were replaced with random labels in \citep{zhang2017understanding}.
\end{remark}

% ---------------------------------------------
\subsection{Generalization Gap Bounds for Specific Probability Distributions}
\label{sub:Specific}

\begin{table}[t]
  \centering
  \caption{Generalization gap upper bounds and sufficient conditions for data length (main terms only) for specific probability distributions. $(q_j)_{j=0}^{+\infty}$ are the probability masses given by $Q$, sorted in descending order.}
  \label{table:SpecificDistributions}
  \vspace{1ex}

  % makecell のデフォルトを「縦中央・横左揃え」に
  \renewcommand\cellalign{cl}

  \begin{tabular}{llll}
    \toprule
    \textbf{Type of Distribution} &
    \textbf{Condition} &
    \textbf{Generalization Gap} &
    \textbf{Sufficient Data Length} \\
    \midrule
    \makecell{Distribution on a\\ finite set} &
    \makecell{$|\calf{Z}| < +\infty$} &
    \makecell{$O\left(\sqrt{|\calf{Z}| \frac{\ln n}{2n}}\right)$} &
    \makecell{$O\left(|\calf{Z}|\ln|\calf{Z}|\cdot \frac{1}{\varepsilon^{2}} \ln \frac{1}{\varepsilon^{2}}\right)$} \\
    \makecell{Exponentially decaying\\ distribution} &
    \makecell{\\ $\exists C >0,$\\ $ r \in (0,1),$\\ $q_j \le C r^{j}$ \\ \ } &
    \makecell{$O\left(\sqrt{ \frac{eC}{\ln \frac{1}{r}} \cdot \frac{(\ln n)^{2}}{2n}} \right)$} &
    \makecell{$O\left(\frac{1}{\ln \frac{1}{r}} \cdot \frac{1}{\varepsilon^{2}} \left(\ln \frac{1}{\varepsilon}\right)^{2} \right)$} \\
    \makecell{Power-law decaying\\ distribution} &
    \makecell{$\exists C>0,$ \\ $ \gamma>1,$\\ $q_j \le C (j+1)^{-\gamma}$} &
    \makecell{$O\left( \sqrt{ \frac{(\ln n)^{2}}{(\gamma-1) n^{\frac{\gamma-1}{\gamma}}} } \right)$} &
    \makecell{$O\left( \left(\frac{\gamma^2}{(\gamma-1)^3} \cdot \frac{1}{\varepsilon^2} \left(\ln \frac{1}{\varepsilon}\right)^2\right)^{\frac{\gamma}{\gamma-1}} \right)$} \\
    \bottomrule
  \end{tabular}
\end{table}

Let's see how generalization gap is suppressed through specific probability distributions.
First, let's look at the relatively trivial case where $\calf{Z}$ is a finite set, and that is the only assumption. In this case, which includes the uniform distribution, the \Renyi{} entropy is finite, so a meaningful generalization gap upper bound can be obtained.
Next, we discuss cases where $\calf{Z}$ may be a countably infinite set.
Theorem~\ref{thm:RenyiGenErr} asserted that the generalization gap becomes smaller if the unevenness of the data distribution is larger. In other words, the faster the tail of the probability distribution decays, the smaller the generalization gap.
Here, we compare the case where the tail of the probability distribution decays exponentially and the case where it decays according to a power law, and see that the generalization gap upper bound is smaller for exponential decay, but the upper bound of Theorem~\ref{thm:RenyiGenErr} is not vacuous, i.e., converges to $0$ as $n \to +\infty$ even for power-law decay.
Phenomena with power-law decaying distributions, such as Zipf's law \citep{Zipf1949}, frequently appear especially in natural languages \citep{Lin2017,Ebeling1995,Ebeling1994,Li1989,Sainburg2019,Takahashi2017,Takahashi2019,Tanaka-Ishii2016}.
Therefore, whether machine learning generalizes for phenomena following these distributions is an important problem.
Due to space limitations, only the main terms are shown in Table~\ref{table:SpecificDistributions}. See Section~\ref{sec:SpecificDetails} for details.

% =========================================================
\section{\Renyi{} Entropy Version of No-free-lunch Theorem}
\label{sec:NFL}

The No-free-lunch theorem in the context of machine learning (e.g., \citep{shalev-shwartz2014understanding}) formulates a certain theoretical limitation of machine learning, especially supervised learning. Specifically, it means that even if information that the input-output relationship is a deterministic function is given, any machine learning algorithm will fail in the worst case regarding the input distribution and input-output relationship if there is not enough training data of a length corresponding to the size of the input data space. The following is a more specific version in \citep{suzuki2025hallucinationsinevitablestatisticallynegligible}.

\begin{theorem}[No-free-lunch theorem]
  \label{thm:NFL}
  Consider a learning problem from a domain set $\calf{X}$ to a codomain set $\calf{Y}$ such that $|\calf{Y}| \ge 1$, i.e., $\calf{Y} \neq \emptyset$.
  For a probability measure $Q$ on $\calf{X}$, a ground truth map $f_0 : \calf{X} \to \calf{Y}$, denote the 0-1 risk of a hypothesis map $f : \calf{X} \to \calf{Y}$ on $Q$ and $f_0$ by $\Risk_{(\ell_\textrm{0-1},\, Q \circ (\idop_{\calf{X}}, f_0))}(f)$,
  which is defined by
  \begin{equation}
    \Risk_{(Q \circ (\idop_{\calf{X}}, f_0)^{-1}, \ell_\textrm{0-1})}(f)
    = \Prob_{X \sim Q} ( f(X) \neq f_0(X) ).
  \end{equation}
  Then, for any map (learning algorithm) $\mathfrak{A} : (\calf{X} \times \calf{Y})^{*} \to (\calf{X} \to \calf{Y})$,
  any nonnegative integer (training data size) $n$ that satisfies $n \le \frac{1}{2} |\calf{X}|$, any finite positive integer $p$ satisfying $1 \le p \le |\calf{Y}|$, and any $\varepsilon \in (0,1)$, there exist a ground truth map $f_0 : \calf{X} \to \calf{Y}$ and a finite subset $\underline{\calf{X}} \subset \calf{X}$ such that $Q = \Uniform(\underline{\calf{X}})$, i.e., the uniform distribution on $\underline{\calf{X}}$, satisfies both the following inequalities:
  \begin{equation}
  \EE_{\BM{Z} \sim (Q \circ (\idop_{\calf{X}}, f_0)^{-1})^{n}}
  \Risk_{(\ell_\textrm{0-1},\, Q \circ (\idop_{\calf{X}},f_0))}
  \left( \mathfrak{A}(\BM{Z}) \right)
  \ge \mu_{\mathrm{err}} := \frac{p-1}{2p},
  \end{equation}
  \begin{equation}
  \Prob_{\BM{Z} \sim (Q \circ (\idop_{\calf{X}}, f_0)^{-1})^{n}}
  \left(
    \Risk_{(\ell_\textrm{0-1},\, Q \circ(\idop_{\calf{X}},f_0))}
    \left( \mathfrak{A}(\BM{Z}) \right)
    \ge \varepsilon
  \right)
  \ge
  \delta := \frac{\mu_{\mathrm{err}} - \varepsilon}{1 - \varepsilon}
  = \frac{p - 1 - 2p\varepsilon}{2p - 2p\varepsilon}.
  \end{equation}
\end{theorem}

\begin{remark}
  We are interested in the cases where $|\calf{Y}| \ge 2$ and we can take $p$ so that $p \ge 2$.
  If $p \ge 2$, then $\mu_{\mathrm{err}} \ge \frac{1}{4}$ and $\delta \ge \frac{1 - 4\varepsilon}{4 - 4\varepsilon}$.
  Moreover, if $\varepsilon = \frac{1}{8}$, then $\delta \ge \frac{1}{7}$.
\end{remark}

The statement (and the original proof in \citep{shalev-shwartz2014understanding}) says that the worst distribution is the uniform distribution, in which the training data size should be at least half of the data space size. It has often been pointed out that the situation where the input distribution is uniform is unlikely to apply to real data, and thus has little implication for real-world machine learning \citep{goldblum2024no,wilson2025deeplearningmysteriousdifferent}.
On the other hand, we have seen in Theorem~\ref{thm:GeneralDataLength} that the sufficient condition about the training data length is almost of the order of the exponential of the \Renyi{} entropy (recall Remark~\ref{rem:DataLengthRenyiRelation}), which can be much smaller than the data space cardinality.
It implies that if we know that \Renyi{} entropy is small, then the original no-free-lunch theorem no longer holds since the uniform distribution is no longer allowed.
Now, we have two questions. (1) Is there a no-free-lunch theorem where the distribution is uneven, or its \Renyi{} entropy has an upper limit? (2) If yes, is it consistent with Theorem~\ref{thm:GeneralDataLength}? The answers are yes for both.

\begin{theorem}[No-free-lunch theorem: the \Renyi{} entropy version]
  \label{thm:RenyiNFL}
  Consider the same setting as in Theorem~\ref{thm:NFL}.
  Specifically, consider a learning problem from a domain set $\calf{X}$ to a codomain set $\calf{Y}$ such that $|\calf{Y}| \ge 1$, i.e., $\calf{Y} \neq \emptyset$.
  For a probability measure $Q$ on $\calf{X}$, a ground truth map $f_0 : \calf{X} \to \calf{Y}$, denote the 0-1 risk of a hypothesis map $f : \calf{X} \to \calf{Y}$ on $Q$ and $f_0$ by $\Risk_{(\ell_\textrm{0-1},\, Q \circ (\idop_{\calf{X}}, f_0))}(f)$,
  which is defined by
  \begin{equation}
    \Risk_{(Q \circ (\idop_{\calf{X}}, f_0)^{-1}, \ell_\textrm{0-1})}(f)
    = \Prob_{X \sim Q} ( f(X) \neq f_0(X) ).
  \end{equation}
  Fix an arbitrary nonnegative integer $n_0$. Then, the following statement holds. For any map (learning algorithm) $\mathfrak{A} : (\calf{X} \times \calf{Y})^{*} \to (\calf{X} \to \calf{Y})$,
  any nonnegative integer (training data size) $n$ that satisfies $n \le n_{0}$, any finite positive integer $p$ satisfying $1 \le p \le |\calf{Y}|$, and any $\varepsilon \in (0,1)$, there exist a ground truth map $f_0 : \calf{X} \to \calf{Y}$ and a distribution $Q$ on $\calf{X}$ satisfying
  \begin{equation}
    \exp\left( H_\alpha(Q) \right) \le 2 n_{0}
    \quad \textrm{for all } \alpha \in [0,1]
  \end{equation}
  such that both the inequalities of Theorem~\ref{thm:NFL} hold; that is, both the following inequalities hold:
  \begin{equation}
  \EE_{\BM{Z} \sim (Q \circ (\idop_{\calf{X}}, f_0)^{-1})^{n}}
  \Risk_{(\ell_\textrm{0-1},\, Q \circ(\idop_{\calf{X}},f_0))}
  \left( \mathfrak{A}(\BM{Z}) \right)
  \ge \mu_{\mathrm{err}} := \frac{p-1}{2p},
  \end{equation}
  \begin{equation}
  \Prob_{\BM{Z} \sim (Q \circ (\idop_{\calf{X}}, f_0)^{-1})^{n}}
  \left(
    \Risk_{(\ell_\textrm{0-1},\, Q \circ (\idop_{\calf{X}},f_0))}
    \left( \mathfrak{A}(\BM{Z}) \right)
    \ge \varepsilon
  \right)
  \ge
  \delta := \frac{\mu_{\mathrm{err}} - \varepsilon}{1 - \varepsilon}
  = \frac{p - 1 - 2p\varepsilon}{2p - 2p\varepsilon}.
  \end{equation}
\end{theorem}

\begin{proof}
Consider $\calf{X}' \subset \calf{X}$ such that $|\calf{X}'| = 2 n_0$. Then $Q = \Uniform(\calf{X}')$ satisfies $\exp(H_\alpha(Q)) = 2 n_0$, and since $\frac{1}{2}|\calf{X}'| = n_0$, we obtain Theorem~\ref{thm:RenyiNFL} by applying Theorem~\ref{thm:NFL} with $\calf{X} = \underline{\calf{X}} = \calf{X}'$.
\end{proof}

\begin{remark}
  Theorem~\ref{thm:RenyiNFL} essentially states that if an upper bound ($2 n_{0}$ in the statement) on $\exp(H_\alpha(Q))$ is given, learning will fail in the worst case if the training data length is not at least half (i.e., $n_0$ in the statement) of that upper bound ($2 n_{0}$). We remark that Theorem~\ref{thm:GeneralDataLength} has stated that the sufficient condition with respect to the training data for a good generalisation was also almost proportional to $\exp(H_\alpha(Q))$, as stated in Remark~\ref{rem:DataLengthRenyiRelation}.
  In this sense, Theorem~\ref{thm:GeneralDataLength} is tight with respect to the dependency on $H_\alpha(Q)$.
  Note that the theorem statement itself does not depend on $\alpha$. This is because the constructed worst case is a uniform distribution, and the \Renyi{} entropy of a uniform distribution does not depend on the order $\alpha$.
\end{remark}

\section{Limitations, discussions, and future work}
\label{sec:Discussion}

\subsection{\Renyi{} Entropy May Diverge}

As stated in Remark~\ref{rem:InfiniteEntropy}, if \Renyi{} entropy diverges, Theorems~\ref{thm:RenyiGenErr} and~\ref{thm:GeneralDataLength} give vacuous upper bounds. However, as also stated in Remark~\ref{rem:InfiniteEntropy}, such cases are pathological, and since the no-free-lunch theorem discussed in Section~\ref{sec:NFL} applies unconditionally, such cases are inherently unlearnable without additional assumptions.

\subsection{Can we explain the success of machine learning by actually measuring \Renyi{} entropy?}

Can we explain why existing deep learning and other large-scale machine learning models are successful by measuring the \Renyi{} entropy in the environments where they succeed, using the theorems of this research? The answer, unfortunately, is \Emph{no} in practical terms. To reliably estimate the \Renyi{} entropy of a probability distribution, a data size that overwhelmingly exceeds the number of elements in the data space is naturally required \citep{jiao2015minimax}. This is equivalent to or greater than the data size sufficient for the success of machine learning, as suggested by Remarks~\ref{rem:ModelIndependent} and~\ref{rem:DataLengthModelIndependent}. In other words, it is difficult to explain the success of actual large-scale machine learning models using the theory of this research.

However, this is not a drawback specific to this research. There are many attempts to explain the success of deep learning by assuming the true hypothesis class, but in applications where large-scale machine learning models are successful, estimating the true hypothesis class is usually more difficult than the success of the machine learning model itself. And, due to the existence of the no-free-lunch theorem, the success of machine learning cannot be explained without making assumptions about the true hypothesis class or the class of distributions. Due to these circumstances, in general, learning theories for large-scale machine learning models should be regarded not as explaining actual applications, but as showing one possible scenario for the future success of large-scale machine learning models. This paper consists only of mathematical results, which is inevitable.

\subsection{This research cannot directly explain the double descent phenomenon}

The phenomenon known as double descent \citep{belkin2019reconciling}, where the generalization gap first increases and then decreases again as the scale of the machine learning model is increased, is known. However, the magnitude of the effect of the double descent phenomenon is known to depend, for example, on the number of training epochs \citep{nakkiran2021deep}, and thus depends on the specific configuration of the learning algorithm. For this reason, the double descent phenomenon cannot be explained in principle within the framework of this paper.

However, experimental results from double descent research also show that when the model scale becomes sufficiently large, the generalization gap is stable with respect to changes in model scale (entering the so-called modern regime). Therefore, for the motivation of this paper, which is to understand the conditions for the success of ultra-large models on large-scale data, double descent is not a direct problem. The test error in the deteriorating part during double descent is also known experimentally to decrease with the number of data points in regions with a certain amount of data or more (e.g., Figure 11 in \citep{nakkiran2021deep}). Although the theory of this paper does not directly explain the double descent phenomenon, it is not inconsistent with related experimental results.

\subsection{On the direction of obtaining stronger conclusions under stronger assumptions in the future}

As already stated in Remarks~\ref{rem:ModelIndependent} and~\ref{rem:DataLengthModelIndependent}, Theorems~\ref{thm:RenyiGenErr} and~\ref{thm:GeneralDataLength} are model-independent. The absence of assumptions about the model is an advantage in terms of wide applicability. On the other hand, as a general principle of theoretical analysis, the fewer assumptions a theorem has, the weaker its conclusion.

As stated in the previous section, it is impossible to know the appropriate class containing the true hypothesis or the appropriate class containing the true distribution in actual applications, and it is also impossible to know the appropriate class of models corresponding to them. Therefore, both creating theories with wide applicability at the cost of weaker conclusions and creating theories that provide strong conclusions at the risk of not being theoretically applicable to actual applications are important, and it is not the case that only one is important. This research belongs to the former category in the sense that it makes no assumptions about the model. On the other hand, the direction of trying to obtain stronger conclusions by also placing some assumptions on the smoothness as a function of the model or its information-theoretic complexity is an interesting avenue for future work.

% =========================================================
\section{Contents in Appendix}

Section~\ref{sec:MainProof} in Appendix provides a comprehensive proof of the main theorem (Theorem~\ref{thm:RenyiGenErr}) and explains how the non-trivial generalization gap upper bound can be obtained on the countable data space setting through the method of type.
Section~\ref{sec:SpecificDetails} in Appendix presents the detail results of what we have introduced in Section~\ref{sub:Specific}. Specifically, it provides generalization gap bounds and sufficient data lengths for scenarios where the data space is finite, or the data-generating distribution exhibits exponential or power-law decay.
Finally, Section~\ref{sec:OtherProofs} in Appendix offers proofs for other supporting theorems and propositions.

% =========================================================
\section{Conclusion}

In this paper, we derived a model-independent generalization gap upper bound and showed that, under only the assumption that the algorithm is symmetric, there exists a probabilistic upper bound on the generalization gap determined by \Renyi{} entropy, which does not depend on the specific construction or scale of the model.
These probabilistic upper bounds directly explain existing experimental results where the behavior of the generalization gap of the same model changes when labels are replaced with a uniform distribution.
We also confirmed that the behavior of generalization gap is affected by the unevenness of the distribution using examples of specific probability distributions.
While the probabilistic upper bounds on generalization gap provide sufficient conditions for training data length, by extending the no-free-lunch theorem to situations where \Renyi{} entropy is known, we also showed that these upper bounds are, in a sense, necessary conditions.

One interesting future work direction is to extend our framework to stochastic symmetric algorithms, including stochastic gradient descent method and its variants, as already discussed in Remark~\ref{rem:FutureStochastic}. Other future directions and limitations are discussed in Appendix. While there is room for extension, the current version of our model-independent generalization gap bounds succussfully justifies the use of even larger machine learning models in the future for real-world problems where the data distribution often deviates significantly from uniform.

% =========================================================
% Bibliography (same .bib file as Typst: main.bib)
% \bibliographystyle{plainnat}
\bibliography{main}

% =========================================================
% Appendix: Typst had `#include "appendix.typ"`;
% here we include the TeX counterpart.
\appendix

% note this is the acknowledgments section which is not visible in draft.
% use unnumbered first level headings for the acknowledgments. all
% acknowledgments go at the end of the paper before the list of references.
% moreover, you are required to declare funding (financial activities supporting
% the submitted work) and competing interests (related financial activities
% outside the submitted work). More information about this disclosure can be
% found at:
% \url{https://neurips.cc/Conferences/2025/PaperInformation/FundingDisclosure}
%
% Do *not* include this section in the anonymized submission, only in the final
% paper. You can use the `ack` environment provided in the style file to
% autmoatically hide this section in the anonymized submission.
%
% We typset reference section header manualy in order to reproduce example
% paper. No special effort is required (a user should not override
% `bibliography-opts` as well).
% \section*{References}
%
% References follow the acknowledgments in the camera-ready paper. Use unnumbered
% first-level heading for the references. Any choice of citation style is
% acceptable as long as you are consistent. It is permissible to reduce the font
% size to `small` (9 point) when listing the references. Note that the Reference
% section does not count towards the page limit.
%
% Optionally include supplemental material (complete proofs, additional
% experiments and plots) in appendix. All such materials *SHOULD be included in
% the main submission*.

\section{Proof of Theorem~\ref{thm:RenyiGenErr}}
\label{sec:MainProof}

Theorem~\ref{thm:RenyiGenErr} is shown by the \Emph{Countable Hypothesis Bound} theorem (e.g., Theorem 7.7 in \citep{shalev-shwartz2014understanding}). First, we look at the Countable Hypothesis Bound theorem. The Countable Hypothesis Bound theorem can also be regarded as a special version of the PAC-Bayes bound \citep{mcallester1999pac}.

\subsection{Countable Hypothesis Bound Theorem}

\begin{theorem}[Countable Hypothesis Bound Theorem]
\label{thm-NonUniform}
Fix a support set $\mathcal{Z}$, a universal hypothesis set $\mathcal{H}_{\mathrm{all}}$, and a loss function $\ell : \mathcal{H}_{\mathrm{all}} \times \mathcal{Z} \to \RR$ defined on them.
Fix a hypothesis set $\mathcal{H} \subset \mathcal{H}_{\mathrm{all}}$ which is an at most countable set, and a surjective map (numbering) $h : \NN \to \mathcal{H}$ from the set of natural numbers to $\mathcal{H}$. Also, let $w : \NN \to [0,1]$ be a sub-probability mass function on the set of natural numbers. That is, $\sum_{j=0}^{+\infty} w(j) \le 1$ holds.
Fix an arbitrary probability distribution $P$ on $\mathcal{Z}$.
If $\BM{Z} \sim P^{n}$, then for any $(\delta,\varepsilon) \in (0,1]^{2}$, with probability at least $1-\delta$, the following holds \Emph{simultaneously for all $j \in \NN$}:
\begin{equation}
  \Risk_{(\ell, P)}(h(j)) - \EmpRisk_{(\ell, \BM{Z})}(h(j))
  < \DI(\ell)\sqrt{\frac{\ln \frac{1}{w(j)\delta}}{2n}}
  = \DI(\ell)\sqrt{\frac{\ln \frac{1}{w(j)} + \ln \frac{1}{\delta}}{2n}}.
\end{equation}
\end{theorem}

\begin{proof}
First, confirm the concentration inequality for a single hypothesis.

\begin{lemma}
\label{lem-OneHypothesisBound}
Fix an arbitrary $h \in \mathcal{H}$. Let $\BM{Z} \sim P^{n}$.
For any $(\delta,\varepsilon) \in (0,1]^{2}$, if
\begin{equation}
  n \ge \frac{(\DI(\ell))^{2}}{2\varepsilon^{2}} \ln \frac{2}{\delta},
\end{equation}
then with probability at least $1-\delta$,
\begin{equation}
  \bigl|\Risk_{(\ell,P)}(h) - \EmpRisk_{(\ell,\BM{Z})}(h)\bigr| \le \varepsilon.
\end{equation}
In other words, for any $\delta \in (0,1]$, with probability at least $1-\delta$,
\begin{equation}
  \Risk_{(\ell,P)}(h) - \EmpRisk_{(\ell,\BM{Z})}(h) \le \DI(\ell)\sqrt{\frac{\ln \frac{1}{\delta}}{2n}}.
\end{equation}
\end{lemma}

\begin{proof}
From Hoeffding's inequality, with probability
$1 - 2 \exp\!\left(- \frac{2 n \varepsilon^{2}}{(\DI(\ell))^{2}}\right)$,
\begin{equation}
  \Risk_{(\ell,P)}(h) - \EmpRisk_{(\ell,\BM{Z})}(h) \le \varepsilon.
\end{equation}
It is clear from here.
\end{proof}

Define the set $A_{j} \subset \mathcal{Z}^{n}$ as
\begin{equation}
  A_{j} := \Bigl\{ \BM{z} \in \mathcal{Z}^{n} \;\Bigm|\; \Risk_{(\ell,P)}(h(j)) - \EmpRisk_{(\ell, \BM{z})}(h(j)) \ge \DI(\ell)\sqrt{\frac{\ln \frac{1}{w(j)\delta}}{2n}} \Bigr\}.
\end{equation}
Applying Lemma~\ref{lem-OneHypothesisBound} with $\delta \leftarrow w(j)\delta$, we have
\begin{equation}
  \Prob_{\BM{Z} \sim P^{n}}(\BM{Z} \in A_{j}) \le w(j)\delta.
\end{equation}
What needs to be proven is
\begin{equation}
  \Prob_{\BM{Z} \sim P^{n}}\left( \forall j \in \NN,\, \BM{Z} \notin A_{j} \right) \ge 1-\delta.
\end{equation}
This holds from the following:
\begin{equation}
  \Prob_{\BM{Z} \sim P^{n}}\left( \forall j,\, \BM{Z} \notin A_{j} \right)
  = 1 - \Prob_{\BM{Z} \sim P^{n}}\left( \BM{Z} \in \bigcup_{j=0}^{+\infty} A_{j} \right)
  \ge 1 - \sum_{j=0}^{+\infty} \Prob_{\BM{Z} \sim P^{n}}(\BM{Z} \in A_{j})
  \ge 1 - \delta,
\end{equation}
where the last inequality used $\sum_{j=0}^{+\infty} w(j) \le 1$.
\end{proof}

\subsection{Proof idea and tricks for Theorem~\ref{thm:RenyiGenErr}}

When applying the Countable Hypothesis Bound theorem, the problem is how to define the sub-probability distribution $w$ on the hypothesis set $\mathcal{H}$. Note that in Theorem~\ref{thm-NonUniform}, $w$ is formally a sub-probability distribution on $\NN$, but if we consider the case where $h$ is a bijection, $w$ can be interpreted as a sub-probability distribution on $\mathcal{H}$. That is, the problem is how to assign a sub-probability mass $w(h)$ to each hypothesis $h \in \mathcal{H}$.

Since we want to obtain a model-independent generalization gap upper bound, we want to assign a sub-probability mass that does not depend on the construction of $h$. Therefore, what we should consider is the sub-probability mass based on the data $\BM{z}$ when the output of the learning algorithm is $h$. Then, when $h$ is output by the learning algorithm $\mathfrak{A}$, we want to choose $w : \mathcal{H} \to [0,1]$ such that $\ln \frac{1}{w(h)}$ is as small as possible.

At this time, considering a general learning algorithm $\mathfrak{A} : \mathcal{Z}^{*} \to \mathcal{H}$,
the effective hypothesis set is
$\{ \mathfrak{A}(\BM{z}) \mid \BM{z} \in \mathcal{Z}^{n} \}$.
In the worst case, if $\BM{z} \ne \BM{z}'$, then $\mathfrak{A}(\BM{z}) \ne \mathfrak{A}(\BM{z}')$ always holds, which is equivalent to $\bigl|\{ \mathfrak{A}(\BM{z}) \mid \BM{z} \in \mathcal{Z}^{n} \}\bigr| = |\mathcal{Z}|^{n}$
if $\mathcal{Z}$ is a finite set.

Suppose for some $h \in \mathcal{H}$ there exists $\BM{z} \in \mathcal{Z}^{n}$ such that $h = \mathfrak{A}(\BM{z})$. If $\BM{Z} \sim Q^{n}$, then
\begin{equation}
  \Prob(h = \mathfrak{A}(\BM{Z})) = Q^{n}(\BM{z}) = \prod_{i=1}^{n} Q(z_{i}).
\end{equation}
In this case, from a fundamental theorem of information theory, no matter how $w$ is defined, the \Emph{expected} value of $\ln \frac{1}{w(\mathfrak{A}(\BM{Z}))}$ cannot be less than $n H_{1}(Q)$, where $Q$ is the distribution that generates each $Z_{i}$. Substituting a function linear in $n$ into the $\ln \frac{1}{w(h)}$ part of the Countable Hypothesis Bound theorem, the upper bound on generalization gap does not converge to $0$ in the limit $n \to +\infty$. Thus, no meaningful upper bound is obtained.

However, if $\mathfrak{A}$ is \Emph{symmetric}, then if $\mathcal{Z}$ is a finite set,
\begin{equation}
  \bigl|\{ \mathfrak{A}(\BM{z}) \mid \BM{z} \in \mathcal{Z}^{n} \}\bigr|
  = \frac{(|\mathcal{Z}| + n - 1)!}{(|\mathcal{Z}| - 1)!\, n!}
  \le (|\mathcal{Z}| + n - 1)^{|\mathcal{Z}|-1}
\end{equation}
is polynomial in $n$, which is much smaller than $|\mathcal{Z}|^{n}$. Therefore, a much larger sub-probability mass can be assigned to each element of $\{ \mathfrak{A}(\BM{z}) \mid \BM{z} \in \mathcal{Z}^{n} \}$ than in the non-symmetric case, and the linearity of $\ln \frac{1}{w(h)}$ with respect to $n$ can be avoided. This is, essentially, the \Emph{method of types} \citep{csiszar1982information}, which has developed information theory. This is the main trick of our non-trivial generalization gap bound. Here, the countability of the data space $\mathcal{Z}$ is actively used.

The specific assignment uses the probability mass with which that hypothesis is actually selected. Below, if necessary, add dummy elements to make $\mathcal{Z}$ a countably infinite set, and let $z : \NN \to \mathcal{Z}$ be a fixed bijection (numbering).

First, convert the symmetric algorithm $\mathfrak{A} : \mathcal{Z}^{*} \to \mathcal{H}$ that returns a hypothesis from a data sequence into an equivalent algorithm $\mathfrak{A}' : \NN^{\NN} \to \mathcal{H}$
that returns a hypothesis from a histogram. Here, $\mathfrak{A}(\BM{z}) = \mathfrak{A}'(\mathrm{Hist}_{\BM{z}})$.
$\mathrm{Hist}_{\BM{z}}$ is the histogram of $\BM{z}$, and if $\BM{n} = (n_{0},n_{1},\dots) = \mathrm{Hist}_{\BM{z}}$, then
\begin{equation}
  n_{j} = \sum_{i=1}^{n} \mathds{1}(z_{i} = z(j)).
\end{equation}

Below, for simplicity, assume that if $\BM{n} \ne \BM{n}'$, then $\mathfrak{A}'(\BM{n}) \ne \mathfrak{A}'(\BM{n}')$. If this does not hold, a larger sub-probability mass can be assigned, so the generalization gap upper bound can be made smaller, and thus we do not need to consider it. In this case, define $w(h)$ as
\begin{equation}
  w(h) := \Prob_{\BM{Z} \sim Q^{n}}\left( \mathfrak{A}'(\mathrm{Hist}_{\BM{Z}}) = h \right).
\end{equation}
Then, $w(h)$ as a distribution is eventually equivalent to the multinomial distribution $\mathrm{Mult}_{(Q,n)}$ induced by an i.i.d.\ discrete distribution with data length $n$ and probability distribution $Q$. Here,
\begin{equation}
  \mathrm{Mult}_{(Q,n)}(n_{0},n_{1},\dots) := \binom{n}{n_{0},n_{1},\dots} \prod_{j=0}^{+\infty} q_{j}^{n_{j}},
\end{equation}
where $q_{j} = Q(z(j))$ for $j \in \NN$.

Henceforth, we evaluate the \Emph{self-entropy} $\ln \frac{1}{\mathrm{Mult}_{(Q,n)}(\BM{n})}$
of the multinomial distribution. In particular, we are interested in a probabilistic upper bound.

\subsection{Stirling's Formula}

What we want to evaluate is a probabilistic non-asymptotic upper bound on the following normalized self-entropy of a histogram $(n_{0},n_{1},\dots)$ on the multinomial distribution induced by $n$ i.i.d.\ data points generated by the discrete distribution $Q$ defined on $\mathcal{Z}$:
\begin{equation}
  \frac{1}{n} I_{(Q,n)}(n_{0},n_{1},\dots)
  := \frac{1}{n} \ln \frac{1}{\mathrm{Mult}_{(Q,n)}(\BM{n})}
  = - \frac{1}{n} \ln \left( \binom{n}{n_{0},n_{1},\dots} \prod_{j=0}^{+\infty} q_{j}^{n_{j}} \right).
  \label{def-SelfEntropy}
\end{equation}
Here, $n_{j}$ corresponds to the number of data points whose value is $z(j)$.

For factorials, the following formula is used.

\begin{lemma}[Robbins' Stirling's Formula {\citep{robbins1955remark}}]
\label{lem-Stirling}
For any $n \in \NN_{>0}$, where $\NN_{>0} := \{1, 2, ...\}$,
there exists
\begin{equation}
  \frac{1}{12 n + 1} < \theta_{n} < \frac{1}{12 n}
\end{equation}
such that
\begin{equation}
  n! = \sqrt{2\pi}\, n^{n + 1/2} e^{-n} e^{\theta_{n}},
  \quad\text{i.e.}\quad
  \ln n!
  = \left(n + \tfrac{1}{2}\right) \ln n - n + \tfrac{1}{2} \ln(2\pi) + \theta_{n}.
\end{equation}
\end{lemma}

Thus, for $(n_{1},\dots,n_{k}) \in \NN_{>0}^{k}$,
\begin{equation}
  (2\pi)^{\frac{1-k}{2}} \frac{n^{n + 1/2}}{\prod_{i=1}^{k} n_{i}^{\,n_{i} + 1/2}} \exp(A)
  \;<\;
  \binom{n}{n_{1},\dots,n_{k}}
  \;<\;
  (2\pi)^{\frac{1-k}{2}} \frac{n^{n + 1/2}}{\prod_{i=1}^{k} n_{i}^{\,n_{i} + 1/2}} \exp(B),
\end{equation}
where
\begin{equation}
  A = \frac{1}{12n+1} - \sum_{j=1}^{k} \frac{1}{12n_{j}},
  \qquad
  B = \frac{1}{12n} - \sum_{j=1}^{k} \frac{1}{12n_{j}+1}.
\end{equation}

\begin{lemma}[Multinomial Coefficient Estimation]
\label{lem-MultiCoeff}
Let $\BM{n} \in \NN_{>0}^{k}$, and let $n = \|\BM{n}\|_{1} := \sum_{j=0}^{k-1} n_{j}$. There exists $\theta_{(n,\BM{n})}$ such that $|\theta_{(n,\BM{n})}| \le \frac{k}{12n}$ and
\begin{equation}
  \ln \binom{n}{n_{1},n_{2},\dots,n_{k}}
  = n H_{1}\left(\frac{\BM{n}}{n}\right)
    + \frac{1}{2}\left( \ln n - \sum_{j=1}^{k} \ln n_{j} \right)
    + \frac{1}{2}(k-1)\ln(2\pi)
    + \theta_{(n,\BM{n})}.
\end{equation}
Here, $H_{1}(\BM{n}/n)$ is the Shannon entropy (i.e., \Renyi{} entropy of order $1$) when the normalized histogram $\BM{n}/n$ is regarded as a discrete probability distribution.
\end{lemma}

\begin{proof}
It can be derived as follows using Lemma~\ref{lem-Stirling}.
\begin{align*}
  \ln \binom{n}{n_{1},\dots,n_{k}}
  &= \left(n + \frac{1}{2}\right) \ln n - \sum_{j=1}^{k} \left(n_{j} + \frac{1}{2}\right) \ln n_{j} - \frac{1}{2}(k-1)\ln(2\pi) + \theta_{n} - \sum_{j=1}^{k} \theta_{n_{j}} \\
  &= n\left(- \sum_{j=1}^{k} \frac{n_{j}}{n} \ln \frac{n_{j}}{n}\right)
     + \frac{1}{2}\left(\ln n - \sum_{j=1}^{k} \ln n_{j}\right)
     + \frac{1}{2}(k-1)\ln(2\pi)
     + \theta_{(n,\BM{n})} \\
  &= n H_{1}\left(\frac{\BM{n}}{n}\right)
     + \frac{1}{2}\left(\ln n - \sum_{j=1}^{k} \ln n_{j}\right)
     + \frac{1}{2}(k-1)\ln(2\pi)
     + \theta_{(n,\BM{n})},
\end{align*}
where $\theta_{(n,\BM{n})} := \theta_{n} - \sum_{i=1}^{k} \theta_{n_{i}}$ satisfies $|\theta_{(n,\BM{n})}| \le \frac{k}{12n}$.
\end{proof}

\subsection{Upper Bound on the Self-Entropy of a Multinomial Distribution using KL Divergence}

\begin{lemma}[Upper Bound on the Self-Entropy of a Multinomial Distribution]
\label{lem-SelfEntropyBoundByKL}
Let $\BM{q} := (q_{0},q_{1},\dots)$, where $q_{j} = Q(z(j))$.
Then, for the self entropy of the multinomial distribution defined by \eqref{def-SelfEntropy}, we have
\begin{equation}
  \frac{1}{n} I_{(Q,n)}(n_{0},n_{1},\dots)
  \le
  D_{\mathrm{KL}}\left( \frac{\BM{n}}{n} \,\bigg\|\, \BM{q} \right)
  + \frac{\|\BM{n}\|_{0} - 1}{2n} \ln n
  + \frac{\|\BM{n}\|_{0}}{2n} \ln(2\pi).
\end{equation}
Here, $D_{\mathrm{KL}}$ is the Kullback--Leibler divergence, and the normalized non-negative sequences $\BM{n}/n$ and $\BM{q}$ are regarded as discrete probability distributions.
\end{lemma}

\begin{proof}
We have
\begin{align*}
  \frac{1}{n} I_{(Q,n)}(n_{0},n_{1},\dots)
  &= - \frac{1}{n} \ln \binom{n}{n_{0},n_{1},\dots} \prod_{j=0}^{+\infty} q_{j}^{\,n_{j}} \\
  &= - \frac{1}{n} \ln \binom{n}{n_{0},n_{1},\dots} - \sum_{j=0}^{+\infty} \frac{n_{j}}{n} \ln q_{j} \\
  &= - \frac{1}{n} \ln \binom{n}{n_{0},n_{1},\dots}
     - \sum_{j=0}^{+\infty} \frac{n_{j}}{n} \ln \frac{q_{j}}{n_{j}/n}
     - \sum_{j=0}^{+\infty} \frac{n_{j}}{n} \ln \frac{n_{j}}{n} \\
  &= - \frac{1}{n} \ln \binom{n}{n_{0},n_{1},\dots}
     + D_{\mathrm{KL}}\left(\frac{\BM{n}}{n} \,\bigg\|\, \BM{q} \right)
     + H_{1}\left(\frac{\BM{n}}{n}\right).
\end{align*}

Let $\|\BM{n}\|_{0} := \sum_{j=0}^{+\infty} \mathds{1}(n_{j} > 0)$, and let $\tilde{\BM{n}} \in \NN^{\|\BM{n}\|_{0}}$ be the sequence formed by collecting only the positive elements of $\BM{n}$ (the order does not affect the discussion). For example, if $\BM{n} = (0,3,4,0,1,0,0,0,\dots)$, then $\tilde{\BM{n}} = (3,4,1)$. In this case,
\begin{equation}
  \binom{n}{n_{0},n_{1},\dots}
  = \binom{n}{\tilde{n}_{0},\tilde{n}_{1},\dots,\tilde{n}_{\|\BM{n}\|_{0}-1}},
  \qquad
  H_{1}\left(\frac{\BM{n}}{n}\right) = H_{1}\left(\frac{\tilde{\BM{n}}}{n}\right).
\end{equation}
From this,
\begin{equation}
  \frac{1}{n} I_{(Q,n)}(n_{0},n_{1},\dots)
  = - \frac{1}{n} \ln \binom{n}{\tilde{n}_{0},\dots,\tilde{n}_{\|\BM{n}\|_{0}-1}}
    + D_{\mathrm{KL}}\left(\frac{\BM{n}}{n} \,\bigg\|\, \BM{q} \right)
    + H_{1}\left(\frac{\tilde{\BM{n}}}{n}\right).
\end{equation}
For the binomial part, using Lemma~\ref{lem-MultiCoeff}, and noting that $H_{1}(\tilde{\BM{n}}/n)$ cancels out, we obtain
\begin{align*}
  \frac{1}{n} I_{(Q,n)}(n_{0},n_{1},\dots)
  &= D_{\mathrm{KL}}\left(\frac{\BM{n}}{n} \,\bigg\|\, \BM{q} \right)
     + \frac{1}{2n}\left(\sum_{j=0}^{\|\BM{n}\|_{0}-1} \ln \tilde{n}_{j} - \ln n\right)
     + \frac{\|\BM{n}\|_{0}-1}{2n} \ln(2\pi)
     + \frac{\theta_{(n,\tilde{\BM{n}})}}{n} \\
  &\le D_{\mathrm{KL}}\left(\frac{\BM{n}}{n} \,\bigg\|\, \BM{q} \right)
     + \frac{\|\BM{n}\|_{0}-1}{2n} \ln n
     + \frac{\|\BM{n}\|_{0}}{2n} \ln(2\pi).
\end{align*}
\end{proof}

\subsection{Uniform-type Counting Bound for KL Distance}

\begin{lemma}[KL Upper Bound]
\label{lem-KLLocalUpperBound}
Let $\BM{q} := (q_{0},q_{1},\dots)$, where $q_{j} = Q(z(j))$. For any $k \in \NN_{>0}$ and $\varepsilon > 0$,
\begin{equation}
  \Prob_{\BM{N} \sim \mathrm{Mult}_{(Q,n)}}\left(
    D_{\mathrm{KL}}\left(\frac{\BM{N}}{\|\BM{N}\|_{1}} \middle\| \BM{q} \right) > \varepsilon
    \;\text{ and }\;
    \|\BM{N}\|_{0} = k
  \right)
  \le (n+1)^{k-1} \exp(-n\varepsilon).
\end{equation}
Here, $D_{\mathrm{KL}}$ is the Kullback--Leibler divergence, and the normalized non-negative sequences $\frac{\BM{N}}{\|\BM{N}\|_{1}}$ and $\BM{q}$ are regarded as discrete probability distributions.
\end{lemma}

\begin{proof}
In this proof, $\mathcal{Z}$ is identified with the set of natural numbers $\NN$, and $Q$ is defined on $\NN$. Recall that we define the frequency $\mathrm{Freq}_{\BM{Z}}$ of $\BM{Z}$ by $\mathrm{Freq}_{\BM{Z}} := \frac{\mathrm{Hist}_{\BM{Z}}}{\|\mathrm{Hist}_{\BM{Z}}\|_{1}}$.

Let $\BM{Z} = (Z_{1},\dots,Z_{n}) \sim Q^{n}$, i.e., $Z_{1},\dots,Z_{n}$ are i.i.d.\ random variables generated by $Q$.
We can construct $\BM{N} \sim \mathrm{Mult}_{(Q,n)}$ by $\BM{N} := \mathrm{Hist}_{\BM{Z}}$, where the histogram $\mathrm{Hist}_{\BM{Z}}$ is $\NN^{\NN}$-valued, whose $j$-th element is
\begin{equation}
  [\mathrm{Hist}_{\BM{Z}}]_{j} = \sum_{i=1}^{n} \mathds{1}(Z_{i} = z(j)).
\end{equation}

Now we have
\begin{equation}
  Q^{n}(\BM{Z}) = \prod_{j=0}^{+\infty} q_{j}^{[\mathrm{Hist}_{\BM{Z}}]_{j}} = \prod_{j=0}^{+\infty} q_{j}^{N_{j}}.
\end{equation}
That is,
\begin{align}
  - \frac{1}{n} \ln Q^{n}(\BM{Z})
  &= - \sum_{j=0}^{+\infty} \frac{N_{j}}{n} \ln q_{j} \notag \\
  &= - \sum_{j=0}^{+\infty} \frac{N_{j}}{n} \ln \frac{q_{j}}{N_{j}/n}
     - \sum_{j=0}^{+\infty} \frac{N_{j}}{n} \ln \frac{N_{j}}{n} \notag \\
  &= D_{\mathrm{KL}}\left( \frac{\BM{N}}{n}\middle\| \BM{q} \right) + H_{1}\left( \frac{\BM{N}}{n} \right)
   = D_{\mathrm{KL}}\left( \frac{\BM{N}}{\|\BM{N}\|_{1}}\middle\| \BM{q} \right) + H_{1}\left( \frac{\BM{N}}{\|\BM{N}\|_{1}} \right).
  \label{eqn-QHD}
\end{align}

Define
\begin{equation}
  \mathcal{P}_{n,k} := \{ \BM{p} \in \NN^{\NN} \mid \|\BM{p}\|_{0} = k \}.
\end{equation}
We have the following, using $\BM{N} = \mathrm{Hist}_{\BM{Z}}$ and $\mathrm{Freq}_{\BM{Z}} := \frac{\BM{N}}{\|\BM{N}\|_{1}}$:
\begin{align*}
  &\Prob_{\BM{N} \sim \mathrm{Mult}_{(Q,n)}}\left(
    D_{\mathrm{KL}} \left(\frac{\BM{N}}{\|\BM{N}\|_{1}} \middle\| \, \BM{q}\right) > \varepsilon \text{ and } \|\BM{N}\|_{0} = k
  \right) \\
  &= \Prob_{\BM{N} \sim \mathrm{Mult}_{(Q,n)}}\left(
    D_{\mathrm{KL}}\left(\frac{\BM{N}}{\|\BM{N}\|_{1}} \middle\| \, \BM{q}\right) > \varepsilon \text{ and } \left\|\frac{\BM{N}}{\|\BM{N}\|_{1}}\right\|_{0} = k
  \right) \\
  &= \Prob_{\BM{Z} \sim Q^{n}}\left(
    D_{\mathrm{KL}} \left(\mathrm{Freq}_{\BM{Z}} \middle\| \, \BM{q}\right) > \varepsilon \text{ and } \|\mathrm{Freq}_{\BM{Z}}\|_{0} = k
  \right) \\
  &= \sum_{\hat{\BM{p}} \in \mathcal{P}_{n,k}} \mathds{1}\left( D_{\mathrm{KL}}(\hat{\BM{p}} \,\|\, \BM{q}) > \varepsilon \right)
     \, \Prob_{\BM{Z} \sim Q^{n}}\left( \mathrm{Freq}_{\BM{Z}} = \hat{\BM{p}} \right) \\
  &= \sum_{\hat{\BM{p}} \in \mathcal{P}_{n,k}} \mathds{1}\left( D_{\mathrm{KL}}(\hat{\BM{p}} \,\|\, \BM{q}) > \varepsilon \right)
     \sum_{\BM{z} \in \{ \BM{z}' \mid \mathrm{Freq}_{\BM{z}'} = \hat{\BM{p}} \}}
     Q^{n}(\BM{z}) \\
  &\le \sum_{\hat{\BM{p}} \in \mathcal{P}_{n,k}} \exp\left( n H_{1}(\hat{\BM{p}}) \right) \exp\left( -n(\varepsilon + H_{1}(\hat{\BM{p}})) \right)
      \\
  &\le (n-1)^{k-1} \exp(-n \varepsilon).
\end{align*}
Here we used the fact that for a fixed type the number of sequences of that type is at most $\exp(n H_{1}(\hat{\BM{p}}))$, and the bound \eqref{eqn-QHD}.
\end{proof}

\begin{lemma}
\label{lem-KLUpperBound}
When $\BM{N} \sim \mathrm{Mult}_{(Q,n)}$, the following holds with probability at least $1-\delta$:
\begin{equation}
  D_{\mathrm{KL}}\left(\frac{\BM{N}}{\|\BM{N}\|_{1}} \middle\|\, Q \right)
  \le \frac{1}{n} \left( (\|\BM{N}\|_{0}-1) \ln(n-1) + \ln \tfrac{1}{\delta} \right).
\end{equation}
\end{lemma}

\begin{proof}
From Lemma~\ref{lem-KLLocalUpperBound},
\begin{equation}
  \delta = (n-1)^{k-1} \exp(-n\varepsilon)
  \iff \varepsilon = \frac{1}{n}\left( (k-1)\ln(n-1) + \ln \tfrac{1}{\delta} \right).
\end{equation}
Therefore, for $\BM{N} \sim \mathrm{Mult}_{(Q,n)}$, with probability at least $1-\delta$,
\begin{equation}
  D_{\mathrm{KL}}\left(\frac{\BM{N}}{\|\BM{N}\|_{1}} \middle\|\, Q \right)
  \le \frac{1}{n}\left( (\|\BM{N}\|_{0}-1)\ln(n-1) + \ln \tfrac{1}{\delta} \right).
\end{equation}
\end{proof}

\subsection{Concentration of the number of distinct symbols $\|\BM{N}\|_{0}$}

\begin{lemma}[Concentration inequality for $\|\BM{N}\|_{0}$]
\label{lem-SymbolConcentration}
When $\BM{N} \sim \mathrm{Mult}_{(Q,n)}$, the following holds with probability at least $1-\delta$:
\begin{equation}
  \bigl| \|\BM{N}\|_{0} - \EE\|\BM{N}\|_{0} \bigr|
  < \sqrt{ \frac{n}{2} \ln \frac{2}{\delta} }.
\end{equation}
\end{lemma}

\begin{proof}
$\BM{N}$ can be constructed using $\BM{X} \sim Q^{n}$ as $\BM{N} = \mathrm{Hist}_{\BM{X}}$.
If $\BM{x}, \BM{x}' \in \NN^{n}$ differ only in the $i$-th component, which are $x_{i}$ and $x_{i}'$ respectively, then
\begin{equation}
  \bigl| \|\mathrm{Hist}_{\BM{x}}\|_{0} - \|\mathrm{Hist}_{\BM{x}'}\|_{0} \bigr| \le 1.
\end{equation}
Therefore, applying McDiarmid's inequality to $\|\mathrm{Hist}_{\BM{x}}\|_{0}$ completes the proof.
\end{proof}

\subsection{Upper bound of the expectation of $\|\BM{N}\|_{0}$ using \Renyi{} entropy}

\begin{lemma}[Upper bound of the expectation of $\|\BM{N}\|_{0}$ using \Renyi{} entropy]
\label{lem-BoundOnExpectedSymbols}
For any $Q$ and $\alpha \in [0,1]$, (assuming $0^{0} = 0$,)
\begin{equation}
  \EE_{\BM{N} \sim \mathrm{Mult}(Q,n)} \bigl[ \|\BM{N}\|_{0} \bigr]
  \le n^{\alpha} \sum_{j=0}^{+\infty} q_{j}^{\alpha}
  = n^{\alpha} \exp\left( (1-\alpha) H_{\alpha}(Q) \right).
\end{equation}
\end{lemma}

\begin{proof}
For $x \in [0,1]$, we have $x \le x^{\alpha}$ if $\alpha \in [0,1]$.
Therefore $y \ge 0 \Rightarrow \min\{1,y\} \le y^{\alpha}$.

\begin{equation}
  \EE_{\BM{N} \sim \mathrm{Mult}(Q,n)} \bigl[ \|\BM{N}\|_{0} \bigr]
  = \sum_{j=0}^{+\infty} \left( 1 - (1 - q_{j})^{n} \right)
  \le \sum_{j=0}^{+\infty} \min\{1, n q_{j}\}
  \le n^{\alpha} \sum_{j=0}^{+\infty} q_{j}^{\alpha}
  = n^{\alpha} \exp\left( (1-\alpha) H_{\alpha}(Q) \right).
\end{equation}
\end{proof}

\subsection{Completion of the proof of Theorem~\ref{thm:RenyiGenErr}}

\begin{proof}[Proof of Theorem~\ref{thm:RenyiGenErr}]
From the countable hypothesis bound theorem (Theorem~\ref{thm-NonUniform}) with
\begin{equation}
  w(h) = \mathrm{Mult}_{(Q,n)}(\mathrm{Hist}_{\BM{Z}})
\end{equation}
where $\BM{Z}$ satisfies $\mathfrak{A}(\BM{Z}) = h$, the following inequality holds with probability $1-\delta_{1}$:
\begin{equation}
  \Risk_{(\ell,Q)}\left( \mathfrak{A}(\BM{Z}) \right)
  - \EmpRisk_{(\ell,\BM{Z})}\left( \mathfrak{A}(\BM{Z}) \right)
  \le
  \DI(\ell)
  \sqrt{
    \frac{
      - \ln \mathrm{Mult}_{(Q,n)}(\mathrm{Hist}_{\BM{Z}}) + \ln \tfrac{1}{\delta_{1}}
    }{2n}
  }.
\end{equation}
Also, from the evaluation of self-entropy (Lemma~\ref{lem-SelfEntropyBoundByKL}) by Stirling's inequality, it always holds that
\begin{equation}
  - \frac{1}{n} \ln \mathrm{Mult}_{(Q,n)}(\mathrm{Hist}_{\BM{Z}})
  \le
  D_{\mathrm{KL}}\left( \frac{\mathrm{Hist}_{\BM{Z}}}{n} \middle\|\, Q \right)
  + \frac{\|\mathrm{Hist}_{\BM{Z}}\|_{0} - 1}{2n} \ln n
  + \frac{\|\mathrm{Hist}_{\BM{Z}}\|_{0}}{2n} \ln(2\pi).
\end{equation}
Also, from the concentration inequality (Lemma~\ref{lem-KLUpperBound}) for $D_{\mathrm{KL}}(\mathrm{Hist}_{\BM{Z}}/n \,\|\, Q)$, with probability $1-\delta_{2}$,
\begin{equation}
  D_{\mathrm{KL}}\left( \mathrm{Hist}_{\BM{Z}}/n \,\|\, Q \right)
  \le \frac{1}{n}\left( (\|\mathrm{Hist}_{\BM{Z}}\|_{0} - 1)\ln(n-1) + \ln \tfrac{1}{\delta_{2}} \right).
\end{equation}
That is, with probability at least $1-\delta_{2}$,
\begin{equation}
  - \frac{1}{n} \ln \mathrm{Mult}_{(Q,n)}(\mathrm{Hist}_{\BM{Z}})
  \le
  \frac{\|\mathrm{Hist}_{\BM{Z}}\|_{0}}{2n}
    \left( 3 \ln n + \ln(2\pi) + \ln \tfrac{1}{\delta_{2}} \right).
\end{equation}
Furthermore, by Lemmas~\ref{lem-SymbolConcentration} and~\ref{lem-BoundOnExpectedSymbols}, for any $\alpha \in [0,1]$, with probability at least $1-\delta_{3}$,
\begin{equation}
  \|\mathrm{Hist}_{\BM{Z}}\|_{0}
  \le
  n^{\alpha} \exp\left( (1-\alpha) H_{\alpha}(Q) \right)
  + \sqrt{\frac{n}{2} \ln \frac{1}{\delta_{3}} }
  = \kappa_{(Q,\alpha)}(n) + \sqrt{\frac{n}{2} \ln \frac{1}{\delta_{3}} }.
\end{equation}
This completes the proof.
\end{proof}

\section{Details of Specific Examples}
\label{sec:SpecificDetails}

\subsection{Case where the data space is a finite set}
\label{sub-FiniteSpace}

If it is known that the data space $\mathcal{Z}$ is a finite set, then at least $H_{0}(Q) = \ln |\mathcal{Z}|$ can be said, so the following holds.

\begin{corollary}[Generalization gap upper bound for a finite set]
\label{cor-FiniteSet}
Fix a universal hypothesis set $\mathcal{H}_{\mathrm{all}}$, a data space $\mathcal{Z}$ which is a finite set, and a loss function $\ell : \mathcal{H}_{\mathrm{all}} \times \mathcal{Z} \to \RR$ defined on their Cartesian product. Define $\DI(\ell)$ as in Theorem~\ref{thm:RenyiGenErr}. Let $\mathfrak{A} : \mathcal{Z}^{*} \to \mathcal{H}_{\mathrm{all}}$ be a symmetric machine learning algorithm in the sense of Definition~\ref{def:SymAlg}.

When $n \in \NN_{>0}$ and $\BM{Z} = (Z_{1},\dots,Z_{n}) \sim Q^{n}$, for any $\delta_{1},\delta_{2},\delta_{3} > 0$, the following holds with probability at least $1-(\delta_{1}+\delta_{2}+\delta_{3})$:
\begin{equation}
  \GenGap_{(\ell,Q,\BM{Z})}\left( \mathfrak{A}(\BM{Z}) \right)
  \le
  \DI(\ell) \sqrt{
    \frac{
      \left( |\mathcal{Z}| + \sqrt{ \tfrac{n}{2} \ln \tfrac{2}{\delta_{3}} } \right)
      \left( 3 \ln n + \ln(2\pi) + \ln \tfrac{1}{\delta_{2}} \right)
      + \ln \tfrac{1}{\delta_{1}}
    }{2n}
  }.
\end{equation}
Alternatively, for any $\delta,\varepsilon > 0$, if
\begin{equation}
  n > \max \Bigl\{ 24 |\mathcal{Z}| \ln |\mathcal{Z}| \cdot \frac{1}{\varepsilon^{2}} \ln \frac{12}{\varepsilon^{2}},\; \omega(\delta,\varepsilon) \Bigr\},
\end{equation}
then, when $\BM{Z} \sim Q^{n}$, with probability at least $1-\delta$,
\begin{equation}
  \GenGap_{(\ell,Q,\BM{Z})}\left( \mathfrak{A}(\BM{Z}) \right) < \DI(\ell)\,\varepsilon.
\end{equation}
\end{corollary}

\begin{remark}
The required data size when $\mathcal{Z}$ is a finite set is $O(|\mathcal{Z}| \ln |\mathcal{Z}|)$, which is the same order as the coupon collector's problem with $|\mathcal{Z}|$ coupons.
\end{remark}

\subsection{Case where the probability distribution decays exponentially}

Corollary~\ref{cor-FiniteSet} holds whenever $\mathcal{Z}$ is a finite set, but it does not use the unevenness of the probability distribution. As a result, the conclusion that a data length of $O(|\mathcal{Z}|\ln|\mathcal{Z}|)$ is sufficient is obtained, but this conclusion is not very interesting in real-world machine learning because $|\mathcal{Z}|$ is large. This section and the next section derive better generalization gap upper bounds by actively using information about the decay rate of the probability distribution.

\begin{corollary}[Generalization gap upper bound for exponentially decaying probability distributions]
\label{prp-ExpDecay}
Fix a universal hypothesis set $\mathcal{H}_{\mathrm{all}}$, a data space $\mathcal{Z}$ which is a countably infinite set, and a loss function $\ell : \mathcal{H}_{\mathrm{all}} \times \mathcal{Z} \to \RR$ defined on their Cartesian product. Define $\DI(\ell)$ as in Theorem~\ref{thm:RenyiGenErr}. Let $\mathfrak{A} : \mathcal{Z}^{*} \to \mathcal{H}_{\mathrm{all}}$ be a symmetric machine learning algorithm in the sense of Definition~\ref{def:SymAlg}.

Assume that the (discrete) probability distribution $Q$ decays exponentially. That is, assume there exist a bijection (numbering) $z : \NN \to \mathcal{Z}$ and $r \in (0,1)$, $C > 0$ such that for $j \in \NN$, $Q(z(j)) \le C r^{j}$.

When $n \in \NN_{>0}$ and $\BM{Z} = (Z_{1},\dots,Z_{n}) \sim Q^{n}$, for any $\delta_{1},\delta_{2},\delta_{3} > 0$, the following holds with probability at least $1-(\delta_{1}+\delta_{2}+\delta_{3})$:
\begin{equation}
  \GenGap_{(\ell,Q,\BM{Z})}\left( \mathfrak{A}(\BM{Z}) \right)
  \le
  \DI(\ell)
  \sqrt{
    \frac{
      \left( \tfrac{eC}{\ln \frac{1}{r}} (\ln n + \ln \tfrac{1}{r}) + \sqrt{ \tfrac{n}{2} \ln \tfrac{2}{\delta_{3}} } \right)
      \left( 3\ln n + \ln(2\pi) + \ln \tfrac{1}{\delta_{2}} \right)
      + \ln \tfrac{1}{\delta_{1}}
    }{2n}
  }.
\end{equation}
Alternatively, for any $\delta,\varepsilon \in (0,1]$, if $n > \omega(\delta,\varepsilon)$ and, letting $\tilde{C} := \max\{C,1\}$,
\begin{align*}
  n > {} & \max \left\{
    \frac{36 e \tilde{C}}{\varepsilon^{2} \ln \frac{1}{r}}
      \left( \ln \frac{36 e \tilde{C}}{\varepsilon^{2} \ln \frac{1}{r}} \right)^{2}, 
    \frac{12 e \tilde{C}}{\varepsilon^{2}}
    \left( 3 + \frac{\ln \frac{6 \pi e^{3}}{\delta}}{\ln \frac{1}{r}} \right)
    \ln \left( \frac{6 e \tilde{C}}{\varepsilon^{2}} \left( 3 + \frac{\ln \frac{6 \pi e^{3}}{\delta}}{\ln \frac{1}{r}} \right) \right), 
    \frac{6 e \tilde{C}}{\varepsilon^{2}} \left( 1 + \frac{1}{\ln \frac{1}{r}} \right) \ln \frac{2 \pi}{\delta}
  \right\},
\end{align*}
then, when $\BM{Z} \sim Q^{n}$, with probability at least $1-\delta$,
\begin{equation}
  \GenGap_{(\ell,Q,\BM{Z})}\left( \mathfrak{A}(\BM{Z}) \right) < \DI(\ell)\,\varepsilon.
\end{equation}
\end{corollary}

\begin{remark}
In Corollary~\ref{prp-ExpDecay}, the main term of the sufficient condition for data length $n$ is
\begin{equation}
  O\!\left(
    \frac{1}{\varepsilon^{2} \ln \frac{1}{r}} \left( \ln \tfrac{1}{\varepsilon^{2} \ln \frac{1}{r}} \right)^{2}
  \right)
\end{equation}
with respect to $\varepsilon$ and $r$. The faster the tail probability decays, i.e., the smaller $r$ is, the smaller the required data length. Also, the order with respect to $\varepsilon$ is roughly $1/\varepsilon^{2}$, which is an order often shown in many learning theories (e.g., global Rademacher complexity).
\end{remark}

\subsection{Case where the probability distribution decays according to a power law}

For example, in natural language, phenomena with power-law decaying distributions, such as Zipf's law \citep{Zipf1949}, frequently appear \citep{Lin2017,Ebeling1995,Ebeling1994,Li1989,Sainburg2019,Takahashi2017,Takahashi2019,Tanaka-Ishii2016}. Therefore, whether machine learning generalizes for phenomena following these distributions is an important problem.

In fact, it can be said that even for power-law decaying probability distributions, generalization is possible, although it requires more training data compared to the case of exponentially decaying distributions, as follows.

\begin{corollary}[Generalization gap upper bound for power-law decaying probability distributions]
\label{prp-PolyDecay}
Fix a universal hypothesis set $\mathcal{H}_{\mathrm{all}}$, a data space $\mathcal{Z}$ which is a countably infinite set, and a loss function $\ell : \mathcal{H}_{\mathrm{all}} \times \mathcal{Z} \to \RR$ defined on their Cartesian product. Define $\DI(\ell)$ as in Theorem~\ref{thm:RenyiGenErr}. Let $\mathfrak{A} : \mathcal{Z}^{*} \to \mathcal{H}_{\mathrm{all}}$ be a symmetric machine learning algorithm in the sense of Definition~\ref{def:SymAlg}.

Assume that the (discrete) probability distribution $Q$ decays according to a power law. That is, assume there exist a bijection (numbering) $z : \NN \to \mathcal{Z}$ and $C>0$, $\gamma > 1$ such that for $j \in \NN$, $Q(z(j)) \le C (j+1)^{-\gamma}$.

When $n \in \NN_{>0}$ and $\BM{Z} = (Z_{1},\dots,Z_{n}) \sim Q^{n}$, for any $\delta_{1},\delta_{2},\delta_{3} > 0$, the following holds with probability at least $1-(\delta_{1}+\delta_{2}+\delta_{3})$:
\begin{equation}
  \GenGap_{(\ell,Q,\BM{Z})}\left( \mathfrak{A}(\BM{Z}) \right)
  \le
  \DI(\ell)
  \sqrt{
    \frac{
      \left( \tfrac{e \tilde{C}}{\gamma-1} n^{\frac{1}{\gamma}} (\ln n + \gamma)
             + \sqrt{ \tfrac{n}{2} \ln \tfrac{2}{\delta_{3}} } \right)
      \left( 3 \ln n + \ln(2\pi) + \ln \tfrac{1}{\delta_{2}} \right)
      + \ln \tfrac{1}{\delta_{1}}
    }{2n}
  },
\end{equation}
where $\tilde{C} := \max\{C,1\}$.

Alternatively, for any $\delta,\varepsilon \in (0,1]$, if $n > \omega(\delta,\varepsilon)$ and
\begin{align*}
  n > \max \Biggl\{ &
    \left( \frac{9 e \tilde{C}}{\varepsilon^{2}(\gamma-1)} \right)^{\frac{\gamma}{\gamma-1}}
    \left(
      \frac{4\gamma}{\gamma-1}
      \Bigl[ \ln \frac{2\gamma}{\gamma-1}
             + \frac{1}{2} \ln \frac{9 e \tilde{C}}{\varepsilon^{2}(\gamma-1)}
      \Bigr]_{+}
    \right)^{\frac{2\gamma}{\gamma-1}},
    \\
    &
    \left(
      \frac{6 e \tilde{C}}{2 \varepsilon^{2}}
      \left( \frac{3\gamma}{\gamma-1} + \frac{1}{\gamma-1} \ln \frac{2\pi}{\delta_{2}} \right)
    \right)^{\frac{\gamma}{\gamma-1}}
    \left(
      \frac{2\gamma}{\gamma-1}
      \bigl[ \ln \frac{\gamma}{\gamma-1}
             + \ln \left(
               \frac{6 e \tilde{C}}{2 \varepsilon^{2}}
               \left( \frac{3\gamma}{\gamma-1} + \frac{1}{\gamma-1} \ln \frac{2\pi}{\delta_{2}} \right)
             \right)
      \bigr]_{+}
    \right)^{\frac{\gamma}{\gamma-1}},
    \\
    &
    \left(
      \frac{6 e \tilde{C} \gamma}{\varepsilon^{2}(\gamma-1)} \ln \frac{2\pi}{\delta_{2}}
    \right)^{\frac{\gamma}{\gamma-1}}
  \Biggr\},
\end{align*}
then, when $\BM{Z} \sim Q^{n}$, with probability at least $1-\delta$,
\begin{equation}
  \GenGap_{(\ell,Q,\BM{Z})}\left( \mathfrak{A}(\BM{Z}) \right) < \DI(\ell)\,\varepsilon.
\end{equation}
\end{corollary}

\begin{remark}
In Corollary~\ref{prp-PolyDecay}, looking at the dependence on $\gamma$ of the part related to $\varepsilon$ in the main term of the sufficient condition for data length $n$, it is
\begin{equation}
  O\!\left(
    \left( \frac{1}{\varepsilon^{2}(\gamma-1)} \right)^{\frac{\gamma}{\gamma-1}}
    \left( \ln \tfrac{1}{\varepsilon^{2}} \right)^{\frac{2\gamma}{\gamma-1}}
  \right).
\end{equation}
The larger $\gamma$ is, the faster the decay. Also, the order with respect to $\varepsilon$ is roughly $\varepsilon^{-2\gamma/(\gamma-1)}$. Noting that $\gamma > 1$, the dependence on $1/\varepsilon$ is worse than in the exponential-decay case (Corollary~\ref{prp-ExpDecay}), but generalization still occurs with finite-length data.
\end{remark}

\section{Proofs (for those other than the main theorem)}
\label{sec:OtherProofs}

The following lemma is repeatedly used in these proofs.

\begin{lemma}
\label{lem-LogPolyInverse}
For $\rho \ge 0$ and $b > 0$, if
\begin{equation}
  n > \left( \frac{2 [\ln \frac{1}{b\rho}]_{+}}{b\rho} \right)^{\frac{1}{\rho}},
\end{equation}
then
\begin{equation}
  \frac{\ln n}{n^{\rho}} < b.
\end{equation}
Here, if $\rho = 0$, then
$\left( \frac{2 [\ln \frac{1}{b\rho}]_{+}}{b\rho} \right)^{\frac{1}{\rho}} = +\infty$, meaning a vacuous statement.

Here, for $x \in \RR$, $[x]_{+} := \max\{x,0\}$ is defined.
\end{lemma}

\begin{proof}
If $\rho = 0$, the lemma makes no assertion, so we do not need to consider it. Below, we consider $\rho \in (0,1]$.

We divide into cases based on the relationship between $b$ and $\rho$.

First, if $b \ge \frac{1}{\rho}$, then $(\ln n) / n^{\rho}$ takes its maximum value $\frac{1}{e\rho}$ in the range $n \in (0,+\infty)$ at $n = \exp(\frac{1}{\rho})$. Therefore, if $b \ge \frac{1}{\rho}$, noting that $[\ln \frac{1}{b\rho}]_{+} = 0$, if
\begin{equation}
  n > \left( \frac{2 [\ln \frac{1}{b\rho}]_{+}}{b\rho} \right)^{\frac{1}{\rho}} (=0),
\end{equation}
then $(\ln n)/n^{\rho} \le \frac{1}{e\rho} < \frac{1}{\rho} \le b$.

If $b \in (0, \frac{1}{\rho})$, noting that $[\ln \frac{1}{b\rho}]_{+} = \ln \frac{1}{b\rho}$, from the assumption
\begin{equation}
  n > \left( \frac{2 \ln \frac{1}{b\rho}}{b\rho} \right)^{\frac{1}{\rho}},
\end{equation}
we have
\begin{equation}
  \frac{\ln n}{n^{\rho}} < \frac{b\rho}{2 \ln \frac{1}{b\rho}} \cdot \frac{1}{\rho} \left( \ln (2 \ln \tfrac{1}{b\rho}) + \ln \tfrac{1}{b\rho} \right).
\end{equation}
Here, generally for $x > 0$, $\ln (2x) < x$, so
\begin{equation}
  \ln (2 \ln \tfrac{1}{b\rho}) + \ln \tfrac{1}{b\rho}
  < 2 \ln \tfrac{1}{b\rho}.
\end{equation}
From this,
\begin{equation}
  \frac{b\rho}{2 \ln \frac{1}{b\rho}} \cdot \frac{1}{\rho} \left( \ln (2 \ln \tfrac{1}{b\rho}) + \ln \tfrac{1}{b\rho} \right)
  < \frac{b\rho}{2 \ln \frac{1}{b\rho}} \cdot \frac{1}{\rho} (2 \ln \tfrac{1}{b\rho})
  = b
\end{equation}
can be said, completing the proof.
\end{proof}

Lemma~\ref{lem-LogPolyInverse} can be easily extended as follows.

\begin{lemma}
\label{lem-GenLogPolyInverse}
For $\lambda > 0$, $\rho \ge 0$, and $b > 0$, if
\begin{equation}
  n > \left(
    \frac{2\lambda [\ln \frac{\lambda}{b^{\frac{1}{\lambda}} \rho}]_{+}}{b^{\frac{1}{\lambda}} \rho}
  \right)^{\frac{\lambda}{\rho}}
  = \frac{1}{b^{\frac{1}{\rho}}} \left(
    \frac{2\lambda}{\rho}
    \bigl[ \ln \tfrac{\lambda}{\rho} + \tfrac{1}{\lambda} \ln \tfrac{1}{b} \bigr]_{+}
  \right)^{\frac{\lambda}{\rho}},
\end{equation}
then
\begin{equation}
  \frac{(\ln n)^{\lambda}}{n^{\rho}} < b.
\end{equation}
Here, if $\rho = 0$, then
$\left( \frac{2 [\ln \frac{1}{b\rho}]_{+}}{b\rho} \right)^{\frac{1}{\rho}} = +\infty$, meaning a vacuous statement.
\end{lemma}

\begin{proof}
Since
\begin{equation}
  \frac{(\ln n)^{\lambda}}{n^{\rho}} < b
  \iff \frac{\ln n}{n^{\frac{\rho}{\lambda}}} < b^{\frac{1}{\lambda}},
\end{equation}
we can apply Lemma~\ref{lem-LogPolyInverse} by setting $\rho \leftarrow \frac{\rho}{\lambda}$ and $b \leftarrow b^{\frac{1}{\lambda}}$.
\end{proof}

\begin{proof}[Proof of Theorem~\ref{thm:GeneralDataLength}]
From Theorem~\ref{thm:RenyiGenErr},
\begin{equation}
  \GenGap_{(\ell,Q,\BM{Z})}\left( \mathfrak{A}(\BM{Z}) \right)
  < \DI(\ell) \sqrt{ A_{1} + A_{2} + A_{3} },
\end{equation}
where
\begin{align*}
  A_{1} &:= \frac{ n^{\alpha} \exp((1-\alpha)H_{\alpha}(Q)) (3 \ln n + \ln(2\pi) + \ln \tfrac{1}{\delta_{2}}) }{2n}, \\
  A_{2} &:= \frac{ \sqrt{\frac{n}{2} \ln \frac{2}{\delta_{3}}} (3 \ln n + \ln(2\pi) + \ln \tfrac{1}{\delta_{2}}) }{2n}, \\
  A_{3} &:= \frac{ \ln \tfrac{1}{\delta_{1}} }{2n}.
\end{align*}
Therefore, it is sufficient to show $A_{1} < \varepsilon^{2}/3$, $A_{2} < \varepsilon^{2}/3$, and $A_{3} < \varepsilon^{2}/3$ all hold.

First, from the assumption,
\begin{equation}
  n > \frac{3}{2\varepsilon^{2}} \ln \frac{1}{\delta_{1}}.
\end{equation}
From this,
\begin{equation}
  A_{3} = \frac{\ln \frac{1}{\delta_{1}}}{2n} < \frac{\varepsilon^{2}}{3}.
\end{equation}

Decompose $A_{1} = A_{1,1} + A_{1,2}$ where
\begin{equation}
  A_{1,1} := \frac{3 \exp((1-\alpha)H_{\alpha}(Q)) n^{\alpha} \ln n}{2n}, \quad
  A_{1,2} := \frac{ \exp((1-\alpha)H_{\alpha}(Q)) n^{\alpha} \ln \frac{2\pi}{\delta_{2}} }{2n}.
\end{equation}
It is sufficient to show $A_{1,1} < \varepsilon^{2}/4$ and $A_{1,2} < \varepsilon^{2}/12$.

First, show $A_{1,1} < \varepsilon^{2}/4$.
Let $y := \varepsilon^{2}/12$. Using Lemma~\ref{lem-LogPolyInverse} with $\rho = 1-\alpha$ and $b = y / \exp((1-\alpha)H_{\alpha}(Q))$, the condition
\begin{equation}
  n >
  \left(
    \frac{
      2 (1-\alpha) H_{\alpha}(Q) [ \ln \frac{1}{y(1-\alpha)} ]_{+}
    }{ y (1-\alpha) } \exp((1-\alpha)H_{\alpha}(Q))
  \right)^{1/(1-\alpha)}
\end{equation}
is sufficient. This is satisfied by the assumption
\begin{equation}
  n > \left(
    \frac{24 H_{\alpha}(Q) \ln \frac{12}{\varepsilon^{2}(1-\alpha)}}{\varepsilon^{2}}
  \right)^{1/(1-\alpha)} \exp(H_{\alpha}(Q)).
\end{equation}

Next, show $A_{1,2} < \varepsilon^{2}/12$. This immediately follows from the assumption
\begin{equation}
  n >
  \left(
    \frac{36 \ln \frac{6\pi}{\delta}}{\varepsilon^{2}}
  \right)^{1/(1-\alpha)} \exp(H_{\alpha}(Q))
  =
  \left(
    \frac{36 \exp((1-\alpha)H_{\alpha}(Q)) \ln \frac{6\pi}{\delta} }{\varepsilon^{2}}
  \right)^{1/(1-\alpha)}
\end{equation}
that
\begin{equation}
  A_{1,2} = \frac{ \exp((1-\alpha)H_{\alpha}(Q)) n^{\alpha} \ln \frac{2\pi}{\delta_{2}} }{2n} < \frac{\varepsilon^{2}}{12}.
\end{equation}
From the above, $A_{1} = A_{1,1} + A_{1,2} < \varepsilon^{2}/4 + \varepsilon^{2}/12 = \varepsilon^{2}/3$.

Decompose $A_{2} = A_{2,1} + A_{2,2}$ where
\begin{equation}
  A_{2,1} := \frac{ 3 \ln n \sqrt{ \frac{n}{2} \ln \tfrac{2}{\delta_{3}} } }{2n},
  \quad
  A_{2,2} := \frac{ \ln \frac{2\pi}{\delta_{2}} \sqrt{ \frac{n}{2} \ln \tfrac{2}{\delta_{3}} } }{2n}.
\end{equation}
It is sufficient to show $A_{2,1} < \varepsilon^{2}/12$ and $A_{2,2} < \varepsilon^{2}/4$.

For $A_{2,1}$, using Lemma~\ref{lem-LogPolyInverse} with $b = \varepsilon^{2}/(9 \sqrt{ 2 \ln \tfrac{2}{\delta_{3}} })$ and $\rho = 1/2$, from the assumption
\begin{equation}
  n > \left(
    \frac{18 \sqrt{ \ln \tfrac{2}{\delta_{3}} } }{\varepsilon^{2}}
    \bigl[ \ln \frac{9 \sqrt{ 2 \ln \tfrac{2}{\delta_{3}} }}{\varepsilon^{2}} \bigr]_{+}
  \right)^{2},
\end{equation}
we can say
\begin{equation}
  \frac{\ln n}{\sqrt{n}} < \frac{ \varepsilon^{2} }{ 9 \sqrt{ 2 \ln \tfrac{2}{\delta_{3}} } }.
\end{equation}
This gives
\begin{equation}
  A_{2,1}
  = \frac{ 3 \ln n \sqrt{ \frac{n}{2} \ln \tfrac{2}{\delta_{3}} } }{2n}
  < \frac{ \varepsilon^{2} }{12}.
\end{equation}

For $A_{2,2}$, from the assumption
\begin{equation}
  n >
  \left(
    \frac{18 \sqrt{ \ln \tfrac{2}{\delta_{3}} } }{\varepsilon^{2}}
    \bigl[ \ln \frac{9 \sqrt{ 2 \ln \tfrac{2}{\delta_{3}} }}{\varepsilon^{2}} \bigr]_{+}
  \right)^{2}
  >
  \frac{8 (\ln \frac{2\pi}{\delta_{2}})^{2} \ln \tfrac{1}{\delta_{3}}}{\varepsilon^{4}},
\end{equation}
we have
\begin{equation}
  A_{2,2}
  = \frac{ \ln \frac{2\pi}{\delta_{2}} \sqrt{ \frac{n}{2} \ln \tfrac{2}{\delta_{3}} } }{2n}
  < \frac{\varepsilon^{2}}{4}.
\end{equation}
Therefore
\begin{equation}
  A_{2} = A_{2,1} + A_{2,2}
  < \frac{\varepsilon^{2}}{12} + \frac{\varepsilon^{2}}{4}
  = \frac{\varepsilon^{2}}{3}.
\end{equation}

From the above, since $A_{1} < \varepsilon^{2}/3$, $A_{2} < \varepsilon^{2}/3$, and $A_{3} < \varepsilon^{2}/3$ are all shown, the proof is complete.
\end{proof}

\begin{proof}[Proof of Corollary~\ref{prp-ExpDecay}]
We treat the case where $Q$ has an exponentially decaying tail.

\paragraph{Case assuming exponential decay.}
Let $z : \NN \to \mathcal{Z}$ be a numbering as in the statement, and for $j \in \NN$, define $q_j := Q(z(j))$.
By the assumption of exponential decay, there exist constants $C>0$ and $r \in (0,1)$ such that $q_j \le C r^j$ for all $j \in \NN$.

If $\alpha \in (0,1]$, then
\[
  \sum_{j=0}^{+\infty} q_j^{\alpha}
  \;\le\;
  C^{\alpha} \sum_{j=0}^{+\infty} (r^j)^{\alpha}
  \;\le\;
  \frac{C^{\alpha}}{1 - r^{\alpha}}.
\]
Allowing the right-hand side to be $+\infty$ when $\alpha=1$, we obtain, for all $\alpha \in [0,1]$,
\[
  \exp\bigl((1-\alpha) H_{\alpha}(Q)\bigr)
  \;=\;
  \sum_{j=0}^{+\infty} q_j^{\alpha}
  \;\le\;
  \frac{C^{\alpha}}{1 - r^{\alpha}}.
\]

Using this, we bound $\kappa_{Q}(n)$:
\[
  \kappa_{Q}(n)
  :=
  \min_{\alpha \in [0,1]} n^{\alpha} \sum_{j=0}^{+\infty} q_j^{\alpha}
  \;\le\;
  \min_{\alpha \in [0,1]} \frac{(C n)^{\alpha}}{1 - r^{\alpha}}.
\]

Define $\tilde{\alpha}$ by
\[
  \frac{1}{1 - r^{\tilde{\alpha}}}
  \;=\;
  \frac{\ln(e n) - \ln r}{-\ln r}
  \;=\;
  \frac{\ln(e n)}{\ln((e n)/r)}.
\]
This is equivalent to
\[
  r^{\tilde{\alpha}} = \frac{\ln(e n)}{\ln((e n)/r)}
  \quad\iff\quad
  \tilde{\alpha}
  =
  \frac{
    \ln\!\left(1 + \dfrac{\ln \frac{1}{r}}{\ln(e n)}\right)
  }{
    \ln \frac{1}{r}
  }.
\]
Since $n \ge 1$ implies $\ln(e n)\ge 1$, we have
\[
  0 = \ln 1 < \tilde{\alpha}
  < \frac{1}{\ln(e n)} \le 1,
\]
where the second inequality follows from the general bound $\ln(1+x) < x$ for $x>0$, applied with $x = \frac{\ln(1/r)}{\ln(e n)}$.

Moreover,
\[
  n^{\tilde{\alpha}}
  =
  n^{-\frac{\ln\left(1 - \frac{\ln r}{\ln(e n)}\right)}{\ln r}}
  \le
  n^{1/\ln(e n)}
  <
  n^{1/\ln n}
  = e.
\]
Since $\tilde{\alpha} \in (0,1)$, we also have $C^{\tilde{\alpha}} \le \tilde{C}$, where $\tilde{C} := \max\{C,1\}$. Hence
\[
  \kappa_{Q}(n)
  \le
  \frac{C^{\tilde{\alpha}} n^{\tilde{\alpha}}}{1 - r^{\tilde{\alpha}}}
  \le
  \frac{e \tilde{C}}{\ln \frac{1}{r}}
  \bigl(\ln n + \ln\tfrac{e}{r}\bigr).
\]
Substituting this bound into Theorem~\ref{thm:RenyiGenErr} yields the first (non-asymptotic) inequality of Corollary~\ref{prp-ExpDecay}.

\paragraph{Sample-size bound.}
For the second claim, we substitute the above bound on $\kappa_{Q}(n)$ into the sufficient condition in Theorem~\ref{thm:GeneralDataLength}. The main term that comes from $\kappa_{Q}(n)$ in Theorem~\ref{thm:RenyiGenErr} is
\[
  \frac{1}{2n}
  \,\kappa_{Q}(n)\,
  \bigl(3 \ln n + \ln(2\pi) + \ln\tfrac{1}{\delta_{2}}\bigr)
  \;\le\;
  \frac{e \tilde{C}}{2 n \ln \frac{1}{r}}
  \bigl(\ln n + \ln\tfrac{e}{r}\bigr)
  \bigl(3 \ln n + \ln(2\pi) + \ln\tfrac{1}{\delta_{2}}\bigr).
\]
To guarantee that the total generalization-gap bound is at most $\DI(\ell)\,\varepsilon$, it suffices to require that this term is at most $\varepsilon^{2}/3$. Thus a sufficient condition is
\[
  \frac{e \tilde{C}}{2 n \ln \frac{1}{r}}
  \bigl(\ln n + \ln\tfrac{e}{r}\bigr)
  \bigl(3 \ln n + \ln(2\pi) + \ln\tfrac{1}{\delta_{2}}\bigr)
  < \frac{\varepsilon^{2}}{3}.
\]

It is sufficient that the following three inequalities holds:
\begin{equation}
  \frac{3 e \tilde{C}}{2 \ln \frac{1}{r}} \cdot \frac{(\ln n)^{2}}{n}
  < \frac{\varepsilon^{2}}{6},
  \quad
  \frac{e \tilde{C}}{2 \ln \frac{1}{r}}
   \bigl(3 \ln\tfrac{e}{r} + \ln\tfrac{2\pi}{\delta_{2}}\bigr)
   \cdot \frac{\ln n}{n}
   < \frac{\varepsilon^{2}}{12},
  \quad
  \frac{e \tilde{C}}{2 n \ln \frac{1}{r}}
   \ln\tfrac{e}{r}\,
   \ln\tfrac{2\pi}{\delta_{2}}
   < \frac{\varepsilon^{2}}{12}.
\end{equation}

Applying Lemma~\ref{lem-GenLogPolyInverse} to the first two terms, and solving the last inequality directly for $n$, we obtain the following sufficient conditions:
\begin{equation}
  n >
  \frac{36 e \tilde{C}}{\varepsilon^{2} \ln \frac{1}{r}}
  \left(
    \ln \frac{36 e \tilde{C}}{\varepsilon^{2} \ln \frac{1}{r}}
  \right)^{2},
  \quad
  n >
  \frac{12 e \tilde{C}}{\varepsilon^{2}}
  \left(
    3 + \frac{\ln \frac{6 \pi e^{3}}{\delta_{2}}}{\ln \frac{1}{r}}
  \right)
  \ln\!\left(
    \frac{6 e \tilde{C}}{\varepsilon^{2}}
    \left(
      3 + \frac{\ln \frac{6 \pi e^{3}}{\delta_{2}}}{\ln \frac{1}{r}}
    \right)
  \right),
  \quad
  n >
  \frac{6 e \tilde{C}}{\varepsilon^{2}}
  \left(
    1 + \frac{1}{\ln \frac{1}{r}}
  \right)
  \ln \frac{2\pi}{\delta_{2}}.
\end{equation}
\end{proof}

\begin{proof}[Proof of Corollary~\ref{prp-PolyDecay}]
We now treat the power-law decay case.

\paragraph{Case assuming power-law decay.}
Assume that there exist constants $C>0$ and $\gamma>1$ such that $q_j := Q(z(j)) \le C (j+1)^{-\gamma}$ for all $j \in \NN$
For $\alpha \in (1/\gamma,1]$, we have
\begin{align*}
  \sum_{j=0}^{+\infty} q_j^{\alpha}
  &\le
  C^{\alpha} \sum_{j=0}^{+\infty} (j+1)^{-\alpha\gamma}
  \\
  &\le
  C^{\alpha}
  + C^{\alpha} \int_{0}^{+\infty} (x+1)^{-\alpha\gamma}\,dx
  \\
  &=
  C^{\alpha}
  + C^{\alpha} \int_{1}^{+\infty} x^{-\alpha\gamma}\,dx
  \\
  &=
  C^{\alpha}
  + \frac{C^{\alpha}}{\alpha\gamma - 1}
  = C^{\alpha} \frac{\alpha\gamma}{\alpha\gamma - 1}.
\end{align*}

For the first inequality in the corollary, we bound
\begin{align*}
  \kappa_{Q}(n)
  &:=
  \inf_{\alpha \in (1/\gamma,1]} n^{\alpha} \sum_{j=0}^{+\infty} q_j^{\alpha}
  \\
  &\le
  \inf_{\alpha \in (1/\gamma,1]}
    n^{\alpha} C^{\alpha}
    \sum_{j=0}^{+\infty} (j+1)^{-\alpha\gamma}
  \\
  &\le
  \inf_{\alpha \in (1/\gamma,1]}
    n^{\alpha} C^{\alpha}
    \frac{\alpha\gamma}{\alpha\gamma - 1}.
\end{align*}
Choose
\[
  \alpha_{n}
  :=
  \frac{1}{\gamma}
  + \left(1 - \frac{1}{\gamma}\right) \frac{1}{\ln(e n)}.
\]
Then $\alpha_{n} \in (1/\gamma,1]$ for $n$ large enough, and a direct calculation gives
\begin{align*}
  n^{\alpha_{n}} C^{\alpha_{n}} \frac{\alpha_{n}\gamma}{\alpha_{n}\gamma - 1}
  &\le
  n^{\frac{1}{\gamma} + (1 - \frac{1}{\gamma}) \frac{1}{\ln(e n)}}
  \,\tilde{C}\,
  \frac{1 + \frac{\gamma - 1}{\ln(e n)}}{\frac{\gamma - 1}{\ln(e n)}}
  \\
  &\le
  \tilde{C}\,
  n^{\frac{1}{\gamma} + \frac{1}{\ln n}}
  \left(
    \frac{\ln n + 1}{\gamma - 1} + 1
  \right)
  \\
  &=
  \frac{e \tilde{C}}{\gamma - 1}\, n^{1/\gamma} (\ln n + \gamma),
\end{align*}
where again $\tilde{C} := \max\{C,1\}$. Therefore,
\[
  \kappa_{Q}(n)
  \le
  \frac{e \tilde{C}}{\gamma - 1}\, n^{1/\gamma} (\ln n + \gamma),
\]
and substituting into Theorem~\ref{thm:RenyiGenErr} yields the first inequality in Corollary~\ref{prp-PolyDecay}.

\paragraph{Sample-size bound.}
For the second inequality, the contribution of $\kappa_{Q}(n)$ in Theorem~\ref{thm:RenyiGenErr} is bounded by
\[
  \frac{1}{2n}
  \,\kappa_{Q}(n)\,
  \bigl(3 \ln n + \ln(2\pi) + \ln\tfrac{1}{\delta_{2}}\bigr)
  \;\le\;
  \frac{e \tilde{C}}{2(\gamma-1)n^{1-1/\gamma}}
  (\ln n + \gamma)
  \bigl(3 \ln n + \ln(2\pi) + \ln\tfrac{1}{\delta_{2}}\bigr).
\]
To ensure that the overall generalization-gap bound is at most $\DI(\ell)\,\varepsilon$, it is sufficient that
\[
  \frac{e \tilde{C}}{2(\gamma-1)n^{1-1/\gamma}}
  (\ln n + \gamma)
  \bigl(3 \ln n + \ln(2\pi) + \ln\tfrac{1}{\delta_{2}}\bigr)
  < \frac{\varepsilon^{2}}{3}.
\]

This is implied by the conjunction of the following three inequalities:
\begin{equation}
  \frac{3 e \tilde{C}}{2(\gamma-1)} \cdot \frac{(\ln n)^{2}}{n^{1-1/\gamma}}
  < \frac{\varepsilon^{2}}{6},
  \quad
  \frac{e \tilde{C}}{2(\gamma-1)}
   \bigl(3\gamma + \ln\tfrac{2\pi}{\delta_{2}}\bigr)
   \cdot \frac{\ln n}{n^{1-1/\gamma}}
   < \frac{\varepsilon^{2}}{12},
  \quad
  \frac{e \tilde{C}}{2(\gamma-1)}
   \gamma \ln\tfrac{2\pi}{\delta_{2}}
   \cdot \frac{1}{n^{1-1/\gamma}}
   < \frac{\varepsilon^{2}}{12}.
\end{equation}

Applying Lemma~\ref{lem-GenLogPolyInverse} to the first two, and solving the third inequality directly for $n$, we obtain the following sufficient conditions:
\begin{align*}
  n &>
  \left(
    \frac{9 e \tilde{C}}{\varepsilon^{2}(\gamma-1)}
  \right)^{\frac{\gamma}{\gamma-1}}
  \left(
    \frac{4\gamma}{\gamma-1}
    \Bigl[
      \ln \frac{2\gamma}{\gamma-1}
      + \frac{1}{2}
        \ln \frac{9 e \tilde{C}}{\varepsilon^{2}(\gamma-1)}
    \Bigr]_{+}
  \right)^{\frac{2\gamma}{\gamma-1}},
  \\
  n &>
  \left(
    \frac{6 e \tilde{C}}{2\varepsilon^{2}}
    \left(
      \frac{3\gamma}{\gamma-1}
      + \frac{1}{\gamma-1} \ln \frac{2\pi}{\delta_{2}}
    \right)
  \right)^{\frac{\gamma}{\gamma-1}}
  \left(
    \frac{2\gamma}{\gamma-1}
    \Bigl[
      \ln \frac{\gamma}{\gamma-1}
      + \ln\!\left(
        \frac{6 e \tilde{C}}{2\varepsilon^{2}}
        \left(
          \frac{3\gamma}{\gamma-1}
          + \frac{1}{\gamma-1} \ln \frac{2\pi}{\delta_{2}}
        \right)
      \right)
    \Bigr]_{+}
  \right)^{\frac{\gamma}{\gamma-1}},
  \\
  n &>
  \left(
    \frac{6 e \tilde{C} \gamma}{\varepsilon^{2}(\gamma-1)}
    \ln \frac{2\pi}{\delta_{2}}
  \right)^{\frac{\gamma}{\gamma-1}}.
\end{align*}
\end{proof}

\begin{proof}[Proof of Theorem~\ref{thm:RenyiCauseGenErr}]
We establish the two claimed inequalities.

For $\kappa^{*}_{(Q')}(n)$, we have
\begin{align*}
  \kappa^{*}_{(Q')}(n)
  &=
  \exp\bigl((1-\alpha'^{*}) H_{\alpha'^{*}}(Q')\bigr)
  n^{\alpha'^{*}}
  \\
  &=
  \exp\bigl((1-\alpha'^{*})\bigl(H_{\alpha'^{*}}(Q') - C\bigr)\bigr)
  n^{\alpha'^{*}}
  \exp\bigl((1-\alpha'^{*}) C\bigr)
  \\
  &\ge
  \min_{\alpha \in [0,1]}
  \Bigl[
    \exp\bigl((1-\alpha)\bigl(H_{\alpha}(Q') - C\bigr)\bigr)
    n^{\alpha}
  \Bigr]
  \exp\bigl((1-\alpha'^{*}) C\bigr)
  \\
  &\ge
  \exp\bigl((1-\alpha'^{*}) C\bigr)
  \kappa^{*}_{(Q)}(n),
\end{align*}
where in the last step we used the assumed relationship between $H_{\alpha}(Q')$ and $H_{\alpha}(Q)$ that appears in the statement of the theorem.

For the second claim, concerning $\max\bigl\{\nu_{(Q',\alpha)}(\varepsilon), \widetilde{\nu}_{(Q',\alpha)}(\delta,\varepsilon)\bigr\}$,
the inequality follows immediately by substituting $H_{\alpha}(Q') \ge H_{\alpha}(Q) + C$ (and the corresponding relation for $Q'$ vs.\ $Q$) into the explicit definitions of $\nu_{(\cdot,\alpha)}$ and $\widetilde{\nu}_{(\cdot,\alpha)}$, and comparing the resulting expressions term by term with those for $Q$.
\end{proof}

\end{document}